\documentclass[10pt, conference, compsocconf]{IEEEtran}
\usepackage{moreverb}

\usepackage[utf8]{inputenc}
\usepackage [T1]{fontenc}
\usepackage{graphicx}
\usepackage{amsmath}
\usepackage{amsfonts}
\usepackage{amssymb}
\usepackage{color}
\usepackage{hyperref}
\usepackage{cite}
\usepackage{algorithm}
\usepackage{algpseudocode}

\def\argmax{\text{argmax}}

\def\pegasosrw{{\sc P2PegasosRW}}
\def\pegasosmu{{\sc P2PegasosMU}}
\def\pegasosum{{\sc P2PegasosUM}}
\def\sgn{\mathop{\rm sgn}\nolimits}

\IEEEoverridecommandlockouts

\begin{document}

\title{Gossip Learning with Linear Models on Fully Distributed Data%
\thanks {
  M.~Jelasity was supported by the Bolyai
  Scholarship of the Hungarian Academy of Sciences.
  This work was partially supported by the Future and Emerging
  Technologies programme FP7-COSI-ICT of the European Commission 
  through project QLectives (grant no.: 231200).}
}

\author{\IEEEauthorblockN{R\'obert Orm\'andi, Istv\'an Heged\H us}
\IEEEauthorblockA{University of Szeged\\
Szeged, Hungary\\
\small\tt \{ormandi,ihegedus\}@inf.u-szeged.hu}
\and
\IEEEauthorblockN{M\'ark Jelasity}
\IEEEauthorblockA{University of Szeged and Hungarian Academy of Sciences\\
Szeged, Hungary\\
\small\tt jelasity@inf.u-szeged.hu}
}

\maketitle

\begin{abstract}
Machine learning over fully distributed data poses an important problem in peer-to-peer (P2P) applications.
In this model we have one data record at each network node,
but without the possibility to move raw data due to privacy considerations.
For example, user profiles, ratings, history, or sensor readings can represent this case.
This problem is difficult, because there is no possibility to learn local models, the system model
offers almost no guarantees for reliability, yet the communication cost needs to be kept low.
Here we propose gossip learning, a generic approach that is based on multiple models taking random walks
over the network in parallel, while applying an online learning algorithm to improve themselves, and
getting combined via ensemble learning methods.
We present an instantiation of this approach for the case of classification with linear models.
Our main contribution is an ensemble learning method which---through the continuous combination
of the models in the network---implements a virtual weighted voting mechanism over an 
exponential number of models at practically no extra cost as compared to independent random walks.
We prove the convergence of the method theoretically, and perform extensive experiments on benchmark
datasets.
Our experimental analysis demonstrates the performance and robustness of the proposed approach.
\end{abstract}

\begin{IEEEkeywords}
P2P; gossip; bagging; online learning; stochastic gradient descent; random walk
\end{IEEEkeywords}

\section{Introduction}
\label{sec:intro}


The main attraction of peer-to-peer (P2P) technology for distributed applications and systems is
acceptable scalability at a low cost (no central servers are needed) and a potential for
privacy preserving solutions, where data never leaves the computer of a user in a raw form.
The label P2P covers a wide range of distributed algorithms that follow a specific system model, in which
there are only minimal assumptions about the reliability of communication and the network components.
A typical P2P system consists of a very large number of nodes (peers) that communicate via message passing.
Messages can be delayed or lost, and peers can join and leave the system at any time.

In recent years, there has been an increasing effort to develop collaborative machine learning
algorithms that can be applied in P2P networks.
This was motivated by the various potential applications such as spam filtering, user profile analysis, 
recommender systems and ranking.
For example, for a P2P platform that offers rich functionality to its users including
spam filtering, personalized search, and recommendation~\cite{tribler,gossple-ebdt10,peerson},
or for P2P approaches
for detecting distributed attack vectors~\cite{p2pwormdetection}, complex predictive models
have to be built based on fully distributed, and often sensitive, data.

An important special case of P2P data processing is fully distributed data, where each node holds
only one data record containing personal data, preferences, ratings, history, local sensor readings, and so on.
Often, these personal data records are the most sensitive ones, so it is essential that we
process them locally.
At the same time, the learning algorithm has to be fully distributed, since the usual approach
of building local models and combining them is not applicable.

Our goal here is to present algorithms for the case of fully distributed data.
The design requirements specific to the P2P aspect are the following.
First, the algorithm has to be extremely \emph{robust}.
Even in extreme failure scenarios it should maintain a reasonable performance.
Second, prediction should be possible at any time in a \emph{local} manner;
that is, all nodes should be able to perform high quality prediction immediately without any extra communication.
Third, the algorithm has to have a \emph{low communication complexity}; both in terms
of the number of messages sent, and the size of these messages as well.
Privacy preservation is also one of our main goals, although in this study we do not analyze this
aspect explicitly.

The gossip learning approach we propose involves models that perform a random walk in the P2P network, and that
are updated each time they visit a node, using the local data record.
There are as many models in the network as the number of nodes.
Any online algorithm can be applied as a learning algorithm that is capable of updating models using a continuous stream
of examples.
Since models perform random walks, all nodes will experience a continuous stream of models
passing through them.
Apart from using these models for prediction directly, nodes can also combine them in various ways using
ensemble learning.

The generic skeleton of gossip learning involves three main components: an implementation of random walk,
an online learning algorithm, and ensemble learning.
In this paper we focus on an instantiation of gossip learning, where the online learning method is a stochastic gradient descent
for linear models.
In addition, nodes do not simply update and then pass on models during the random walk, but they also combine
these models in the process.
This implements a distributed ``virtual'' ensemble learning method similar
to bagging, in which we in effect calculate a weighted voting over an exponentially increasing number of
linear models.

Our specific contributions include the following: (1) we propose gossip learning, a novel and generic approach for P2P learning on
fully distributed data, which can be instantiated in various different ways; (2)
we introduce a novel, efficient distributed ensemble learning method for linear models that virtually combines an exponentially
increasing number of linear models; and (3) we provide a theoretical and empirical analysis of the convergence
properties of the method in various scenarios. 

The outline of the paper is as follows.
Section~\ref{sec:fully} elaborates on the fully distributed data model.
Section~\ref{sec:related} summarizes related work and the background concepts.
In Section~\ref{sec:idea} we describe our generic approach and a naive algorithm as an example.
Section~\ref{sec:bagavg} presents the core algorithmic contributions of the paper along with
a theoretical discussion, while Section~\ref{sec:exp} contains an experimental analysis.
Section~\ref{sec:concl} concludes the paper.

This paper is a significantly extended and improved version of our previous work~\cite{ormandi2011asynchronP2PDM}.

\section{Fully Distributed Data}
\label{sec:fully}

Our focus is on fully distributed data, where each node in the network has a single feature vector, that cannot
be moved to a server or to other nodes.
Since this model is not usual in the data mining community, we elaborate on the motivation and the implications of
the model.

In the distributed computing literature the fully distributed data model is typical.
In the past decade, several algorithms have been proposed to calculate distributed aggregation
queries over fully distributed data, such as the average, the maximum, and the network size (e.g., ~\cite{astrolabe,JMB05,boyd05}).
Here, the assumpion is that every node stores only a single record, for example, a sensor reading.
The motivation for not collecting raw data but processing it in place is mainly to achieve
robustness and adaptivity through not relying on any central servers.
In some systems, like in sensor networks or mobile ad hoc networks, the physical constraints
on communication also prevent the collection of the data.

An additional motivation for not moving data is \emph{privacy preservation}, where local data
is not revealed in its raw form, even if the computing infrastructure made it possible.
This is especially important in smart phone applications~\cite{pentland12,mobiscope07,mobilesensing10} and in 
P2P social networking~\cite{diaspora}, where the key motivation
is giving the user full control over personal data.
In these applications it is also common for a user to contribute only a single record,
for example, a personal profile, a search history, or a sensor reading by a smart phone.

Clearly, in P2P smart phone applications and P2P social networks, there is a need for more complex aggregation queries, and ultimately,
for data models, to support features such as recommendations and spam filtering, and to make the system
more robust with the help of, for example, distributed intruder detection.
In other fully distributed systems data models are also important for monitoring and control.
Motivated by the emerging need for building complex data models over fully distributed data in
different systems, we work with the abstraction of fully distributed data, and we aim at proposing
generic algorithms that are applicable in all compatible systems.

In the fully distributed model, the requirements of an algorithm also differ from those
of parallel data mining algorithms, and even from previous work on P2P data mining.
Here, the decisive factor is the cost of message passing.
Besides, the number of messages each node is allowed to send in a given time window is limited,
so computation that is performed locally has a cost that is typically negligible when compared to communication delays.
For this reason prediction performance has to be investigated \emph{as a function of the number of messages sent},
as opposed to wall clock time.
Since communication is crucially important, evaluating robustness to
communication failures, such as message delay and message loss, also gets a large emphasis.

The approach we present here is applicable successfully also when each node stores many records (and not only one);
but its advantages to known approaches to P2P data mining become less significant, since communication plays a smaller
role when local data is already usable to build reasonably good models.
In the following we focus on the fully distributed model.

\section{Background and Related Work}
\label{sec:related}

We organize the discussion of the background of our work along the generic model
components outlined in the Introduction and explained in Section~\ref{sec:idea}:
online learning, ensemble learning, and
peer sampling. We also discuss related work in P2P data mining. 
Here we do not consider parallel data mining algorithms. This field has a large literature,
but the rather different underlying system model means it is of little relevance to us here.

\paragraph{Online Learning.}
The basic problem of {\em supervised binary classification} can be defined as follows.
Let us assume that we are given a labeled database in the form of pairs of feature vectors
and their correct classification, i.e.\ $(x_1,y_1),\ldots,(x_n,y_n)$, where $x_i\in\mathbb{R}^d$,
and $y_i\in \{-1,1\}$.
The constant $d$ is the {\em dimension} of the problem (the number of features).
We are looking for a {\em model} $f: \mathbb{R}^d\rightarrow \{-1,1\}$ that correctly classifies
the available feature vectors, and that can also {\em generalize} well; that is,
which can classify unseen examples too.
For testing purposes, the available data is often partitioned into a {\em training set} and
a {\em test set}, the latter being used only for testing candidate models.

Supervised learning can be thought of as an optimization problem, where we
want to maximize prediction performance, which can be measured via, for example, the number
of feature vectors that are classified correctly over the training set.
The search space of this problem consists of the set of possible models (the
{\em hypothesis space}) and each method also defines a specific search algorithm (often called
the {\em training algorithm}) that eventually selects one model from this space.

Training algorithms
that iterate over available training data, or process a continuous stream of data records, 
and evolve a model by updating it for each individual data record according to some update rule
are called {\em online learning algorithms}. 
Gossip learning relies on this type of learning algorithms.
Ma et al.\ provide a nice summary of online learning for large scale data~\cite{ma-icml09}.

{\em Stochastic gradient search}~\cite{bottou-nips07tut,bottou-lecun-2004} is 
a generic algorithmic family for implementing online learning methods.
Without going into too much detail, the basic idea is that we iterate over
the training examples in a random order repeatedly, and for each training example,
we calculate the gradient of the error function (which describes classification error),
and modify the model along this gradient to reduce the error on this particular example.
At the same time, the step size along the gradient is gradually reduced.
In many instantiations of the method, it can be proven that the
converged model minimizes the {\em sum} of the errors over the examples~\cite{duda-hart-stork-book}.

Let us now turn to support vector machines (SVM), the learning algorithm we apply in this paper~\cite{svm-book}.
In its simplest form, the SVM approach works with the space of linear models to solve the binary
classification problem.
Assuming a $d$ dimensional problem, we want to find a $d-1$ dimensional separating hyperplane
that maximizes the {\em margin} that separates examples of the two class.
The margin is defined by the hyperplane as the sum of the minimal perpendicular distances from both classes.

Equation \eqref{eq:primal} states a variant of the formal SVM optimization problem, where $w\in\mathbb{R}^d$ and $b\in\mathbb{R}$ are the
parameters of model, namely the norm of the separating hyper-plane and the bias parameters, respectively. Furthermore, $\xi_i$ is the 
slack variable of the $i$th sample, which can be interpreted as the amount of misclassification error of the $i$th sample, and $C$ is a
trade-off parameter between generalization and error minimization.
\begin{equation}
\label{eq:primal}
\begin{split}
\min_{w,b,\xi_i}&\quad \frac{1}{2}\|w\|^{2}+C\sum_{i=1}^{n}\xi_i \\
\text{s.t.}&\quad y_i(w^T x_i + b)\geq 1-\xi_i\quad\mbox{and} \\
&\quad \xi_i\geq 0\quad (\forall i:\ 1\leq i\leq n)
\end{split}
\end{equation}

The Pegasos algorithm is an SVM training algorithm, based on a stochastic gradient descent approach~\cite{pegasos10}.
It directly optimizes a form of the above defined, so-called primal optimization task.
We will use the Pegasos algorithm as a basis for our distributed method.
In this primal form, the desired model $w$ is explicitly represented, and is evaluated directly over
the training examples.

Since in the context of SVM learning this is an unusual approach, let us take a closer look
at why we decided to work in the primal formulation.
The standard SVM algorithms solve the dual problem instead of the primal form~\cite{svm-book}.
The dual form is
\vspace{-1mm}
\begin{equation}
\label{eq:dual}
\begin{split}
\max_{\alpha}&\quad \sum_{i=1}^n\alpha_{i} -\frac{1}{2}\sum_{i,j=1}^{n}\alpha_{i}y_{i}\alpha_{j}y_{j}x_i^Tx_j \\
\text{s.t.}&\quad \sum_{i=1}^{n}\alpha_{i}y_{i} = 0 \quad\mbox{and} \\
&\quad 0\leq\alpha_{i}\leq C\quad (\forall i:\ 1\leq i\leq n),
\end{split}
\end{equation}
where the variables $\alpha_i$ are the Lagrangian variables.
The Lagrangian variables can be interpreted as the weights of the training samples,
which specify how important the corresponding sample is from the point of view of the model.

The primal and dual formalizations are equivalent, both in terms of theoretical time complexity and the optimal solution.
Solving the dual problem has some advantages; most importantly, one can take full advantage of the kernel-based
extensions (which we have not discussed here) that introduce nonlinearity into the approach.
However,
methods that deal with the dual form require frequent access to the entire database to update
$\alpha_i$, which is unfeasible in our system model.
Besides, the number of variables $\alpha_i$ equals the number of training samples, which could be orders of magnitude larger
than the dimension of the primal problem, $d$.
Finally, there are indications that applying the primal form can achieve a better generalization on some databases~\cite{primal-svm07}.


\paragraph{Ensemble Learning.}
Most distributed large scale algorithms apply some form of ensemble learning to combine
models learned over different samples of the training data.
Rokach presents a survey of ensemble learning methods~\cite{ensemble-learning10}.
We apply a method for combining the models in the network that is related to both bagging~\cite{bagging-breiman96} and
``pasting small votes''~\cite{breiman-pasting99}:
when the models start their random
walk, initially they are based on non-overlapping small subsets of the training data due to the
large scale of the system (the key idea
behind pasting small votes) and as time goes by, the sample sets grow,
approaching the case of bagging (although the samples that belong to 
different models will not be completely independent in our case).

\paragraph{Peer Sampling in Distributed Systems.}
The sampling probability for each data record is defined by peer sampling algorithms that
are used to implement the random walk.
Here we apply uniform sampling.
A set of approaches to implement uniform sampling in a P2P network
apply random walks themselves over a fixed overlay network, in such
a way that the corresponding Markov-chain has a uniform limiting
distribution~\cite{randpeerpodc04,stutzbach-ton09,hall-europar09}.
In our algorithm, we apply gossip-based peer sampling~\cite{JVGKvS07} where
peers periodically exchange small random subsets of addresses, thereby providing a local
random sample of the addresses at each point in time at each node.
The advantage of gossip-based sampling in our setting is that samples are available locally and
without delay.
Furthermore, the messages related to the peer sampling algorithm
can piggyback the random walks of the models, thereby avoiding any overheads in
terms of message complexity.

\paragraph{P2P Learning.}
In the area of P2P computing, a large number of fully distributed algorithms are known
for calculating global functions over fully distributed data, generally referred to
as aggregation algorithms.
The literature of this field is vast, we mention only two examples: Astrolabe~\cite{astrolabe} and gossip-based
averaging~\cite{JMB05}.
These algorithms are simple and robust, but are capable of calculating only simple
functions such as the average.
Nevertheless, these simple functions can serve as key components
for more sophisticated methods, such as the EM algorithm~\cite{newscastem},
unsupervised learners~\cite{p2pdocclust} or the collaborative filtering based recommender
algorithms~\cite{OHJ10,bakker09,p2prec,Tveit-2001}.
However, here we seek to provide a rather generic approach that covers
a wide range of machine learning models, while maintaining a similar robustness and
simplicity.

In the past few years there has been an increasing number of proposals for P2P machine learning
algorithms as well, like those in~\cite{luo-kdd07,ang-ecml09,ang-ecml10,ang-p2pcascade08,kargupta-p2pdm06,siersdorfer06,gadgetsvm09}.
The usual assumption in these studies is that a peer has a subset of the training data on which
a model can be learned locally.
After learning the local models, algorithms either aggregate the
models to allow each peer to perform local prediction, or they assume that prediction
is performed in a distributed way.
Clearly, distributed prediction is a lot more expensive than local prediction; however,
model aggregation is not needed, and there is more flexibility in the case of changing data.
In our approach we adopt the fully distributed model, where each node holds only one data record.
In this case we cannot talk about local learning: every aspect of the learning algorithm is inherently
distributed.
Since we assume that data cannot be moved, the models need to visit data instead.
In a setting like this, the main problem we need to solve is to efficiently aggregate the various
models that evolve slowly in the system so as to speed up the convergence of prediction performance.

To the best of our knowledge there is no other learning approach designed to work in our fully asynchronous and unreliable
message passing model, and which is capable of producing a large array of state-of-the-art models.

\section{Gossip Learning: the Basic Idea}
\label{sec:idea}

\begin{algorithm}[t]
\caption{Gossip Learning Scheme}
\label{alg:gossiplearn}
\begin{minipage}{60mm}
\begin{algorithmic}[1]
\State initModel()
\Loop
\State $\mbox{wait}(\Delta)$
\State $p \gets$ selectPeer()
\label{line:selectp}
\State send modelCache.freshest() to $p$
\EndLoop
\algstore{gossip}
\end{algorithmic}
\end{minipage}
\hfil
\begin{minipage}{8cm}
\begin{algorithmic}[1]
\algrestore{gossip}
\Procedure{onReceiveModel}{$m$}
\State modelCache.add(createModel($m$, lastModel))
\State lastModel $\gets m$
\EndProcedure
\end{algorithmic}
\end{minipage}
\end{algorithm}

Algorithm~\ref{alg:gossiplearn} provides the skeleton of the gossip learning framework.
The same algorithm is run at each node in the network.
The algorithm consists of an active loop of periodic activity, and a method to handle incoming
models.
Based on every incoming model a new model is created potentially combining it with the previous incoming
model.
This newly created model is stored in a cache of a fixed size.
When the cache is full, the model stored for the longest time is replaced by the newly added model.
The cache provides a pool of recent models that can be used to implement, for example, voting based
prediction.
We discuss this possibility in Section~\ref{sec:exp}.
In the active loop the freshest model (the model added to the cache most recently) is sent to a random peer.

We make no assumptions about either the synchrony of the loops at the different nodes or the
reliability of the messages.
We do assume that the length of the period of the loop $\Delta$ is the same at all nodes.
However, during the evaluations $\Delta$ was modeled as a normally distributed random variable
with parameters $\mu=\Delta$ and $\sigma^2=\Delta/10$.
For simplicity, here we assume that the active loop is initiated at the same time
at all the nodes, and we do not consider any stopping criteria, so the loop runs indefinitely.
The assumption about the synchronized start allows us to focus on the convergence properties of the algorithm,
but it is not a crucial requirement in practical applications.
In fact, randomly restarted loops actually help in following drifting concepts and changing data, which is the
subject of our ongoing work.

The algorithm contains abstract methods that can be implemented in different ways to obtain
a concrete learning algorithm.
The main placeholders are {\sc selectPeer} and {\sc createModel}.
Method {\sc selectPeer} is the interface for the peer sampling service, as described in Section~\ref{sec:related}.
Here we use the {\sc NewsCast} algorithm~\cite{JVGKvS07}, which is a gossip-based implementation
of peer sampling.
We do not discuss {\sc NewsCast} here in detail,
all we assume is that {\sc selectPeer()} provides a \emph{uniform random sample}
of the peers without creating \emph{any extra messages} in the network, given that {\sc NewsCast} gossip messages
(that contain only a few dozen network addresses) can piggyback gossip learning messages.

The core of the approach is {\sc createModel}.
Its task is to create a new updated model based on locally available information---the two models
received most recently, and the local single training data record---to be sent on to a random peer.
Algorithm~\ref{alg:createmodel} lists three implementations that are still abstract.
They represent those three possible ways of breaking down the task that we will study in this paper.

\begin{algorithm}[t]
\caption{{\sc createModel}: three implementations}
\label{alg:createmodel}
\begin{minipage}{7cm}
\begin{algorithmic}[1]
\Procedure{createModelRW}{$m_1,m_2$}
\State \Return update($m_1$)
\EndProcedure
\\
\Procedure{createModelMU}{$m_1,m_2$}
\State \Return update(merge($m_1,m_2$))
\EndProcedure
\algstore{createmodel}
\end{algorithmic}
\end{minipage}
\hfil
\begin{minipage}{7cm}
\begin{algorithmic}[1]
\algrestore{createmodel}
\Procedure{createModelUM}{$m_1,m_2$}
\State \Return merge(update($m_1$),update($m_2$))
\EndProcedure
\end{algorithmic}
\end{minipage}
\end{algorithm}

The abstract method {\sc update} represents the online learning algorithm---the second main component of our framework besides
peer sampling---that updates the model based on one example (the local
example of the node).
Procedure {\sc createModelRW} implements the case where models independently perform random walks over the network.
We will use this algorithm as a baseline.

The remaining two variants apply a method called {\sc merge}, either before the update (MU) or after it (UM).
Method {\sc merge} helps implement the third component: ensemble learning.
A \emph{completely impractical} example for an implementation of {\sc merge} is the case where the model
space consists of all the sets of basic models of a certain type.
Then {\sc merge} can simply merge the two input sets, {\sc update} can update all the models in the set, and
prediction can be implemented via, for example, majority voting (for classification) or averaging the predictions (for regression).
With this implementation, all nodes would collect an exponentially increasing set of models, allowing for a
much better prediction after a much shorter learning time in general than based on a single
model~\cite{bagging-breiman96,breiman-pasting99}, although the learning
history for the members of the set would not be completely independent.

This implementation is of course impractical because the size of the messages in each cycle of the main loop
would increase exponentially.
Our main contribution is to discuss and analyze a special case: linear models.
For linear models we will propose an algorithm where the message size can be
kept \emph{constant}, while producing the same (or similar) behavior as the impractical implementation above.
The subtle difference between
the MU and UM versions will also be discussed.

Let us close this section with a brief analysis of the cost of the algorithm in terms of computation
and communication.
As of communication: each node in the network sends exactly one message in each $\Delta$ time units.
The size of the message depends on the selected hypothesis space; normally it contains the parameters
of a single model.
In addition, the message also contains a small constant number of network addresses as defined by the {\sc NewsCast}
protocol (typically around 20).
The computational cost is one or two update steps in each $\Delta$ time units for the UM or the MU variants, respectively.
The exact cost of this step depends on the selected online learner.

\section{Merging Linear Models through Averaging}
\label{sec:bagavg}





The key observation we make is that in a linear hypothesis space,
in certain cases voting-based prediction is equivalent
to a single prediction by the {\em average} of the models that participate in the voting.
Furthermore, updating a set of linear models and then averaging them is sometimes equivalent
to averaging the models first, and then updating the resulting single model.
These observations are valid in a strict sense only in special circumstances.
However, our intuition is that even if this key observation holds only in
a heuristic sense, it still provides a valid heuristic explanation of the behavior
of the resulting averaging-based merging approach.

In the following we first give an example of a case where there is a strict
equivalence of averaging and voting to illustrate the concept, and subsequently we
discuss and analyze a practical and competitive algorithm, where the correspondence of voting
and averaging is only heuristic in nature.

\subsection{The Adaline Perceptron}
We consider here the Adaline perceptron~\cite{adaline-widrow60} that arguably has
one of the simplest update rules due to its linear activation function.
Without loss of generality, we ignore the bias term.
The error function to be optimized is defined as
\begin{equation}
\label{eq:percErrFunc}
	E_{x}(w) = \frac{1}{2}(y - \langle w,x \rangle)^{2}
\end{equation}
where $w$ is the linear model, and $(x,y)$ is a training example ($x,w\in\mathbb{R}^n$,
$y\in\{-1,1\})$.
The gradient at $w$ for $x$ is given by 
\begin{equation}
\label{eq:percGrad}
	\nabla_{w} = \frac{\partial E_{x}(w)}{\partial w} = -(y - \langle w,x \rangle)x
\end{equation}
that defines the learning rule for $(x,y)$ by
\begin{equation}
\label{eq:percLearnFunc}
	w^{(k+1)} = w^{(k)} + \eta (y - \langle w^{(k)},x \rangle)x,
\end{equation}
where $\eta$ is the learning rate. In this case it is a constant.

Now, let us assume that we are given a set of models $w_1,\ldots,w_m$, and let us define
$\bar w = (w_1+\ldots +w_m)/m$.
In the case of a regression problem, the prediction for a given point $x$ and model $w$
is $\langle w,x\rangle$.
It is not hard to see that
\begin{equation}
h(x)=\langle\bar w,x\rangle=\frac{1}{m}\langle\sum_{i=0}^{m}w_i,x\rangle=\frac{1}{m}\sum_{i=0}^{m}\langle w_i,x\rangle,
\end{equation}
which means that the voting-based prediction is equivalent to prediction based
on the average model.

In the case of classification, the equivalence does not hold for all voting mechanisms.
But it is easy to verify that in the case of a weighted voting approach, where vote
weights are given by $|\langle w,x\rangle|$, and the votes themselves are given
by $\sgn \langle w,x\rangle$, the same equivalence holds:
\begin{equation}
\label{eq:percBagging}
\begin{split}
h(x) & = \sgn(\frac{1}{m}\sum_{i=1}^{m} |\langle w,x\rangle|\sgn \langle w,x\rangle)= \\
& = \sgn(\frac{1}{m}\sum_{i=1}^{m}{\langle w_{i},x} \rangle) = \sgn\langle\bar w,x \rangle.
\end{split}
\end{equation}
A similar approach to this weighted voting mechanism has been shown to improve the
performance of simple vote counting~\cite{vote99}.
Our preliminary experiments also support this.

In a very similar manner, it can be shown that updating $\bar w$ using an example
$(x,y)$ is equivalent to updating all the individual models $w_1,\ldots,w_m$ and then
taking the average:
\begin{equation}
\label{eq:percLearnAvg}
	\bar w + \eta (y - \langle \bar w,x \rangle)x =
	\frac{1}{m}\sum_{i=1}^{m}w_{i} + \eta (y - \langle w_{i},x\rangle)x.
\end{equation}

The above properties lead to a rather important observation.
If we implement our gossip learning skeleton using Adaline, as shown in Algorithm~\ref{alg:p2pegasos},
then the resulting algorithm behaves exactly as if all the models were simply stored and then
forwarded, resulting in an exponentially increasing number of models contained in each message,
as described in Section~\ref{sec:idea}.
That is, averaging effectively reduces the exponential message complexity to transmitting a {\em single} model
in each cycle independently of time, yet we enjoy the benefits of the aggressive, but impractical
approach of simply replicating all the models and using voting over them for prediction.

\begin{algorithm}[tb]
\caption{Pegasos and Adaline updates, initialization, and merging}
\label{alg:p2pegasos}
\begin{minipage}{65mm}
\begin{algorithmic}[1]
\Procedure{updatePegasos}{$m$}
\State $m.t \gets m.t+1$
\State $\eta \gets 1/(\lambda\cdot m.t)$
\If{$y \left\langle m.w, x \right\rangle < 1$}
\State $m.w \gets (1-\eta\lambda)m.w + \eta y x$
\label{line:misclass}
\Else
\State $m.w \gets (1-\eta\lambda)m.w$
\EndIf
\State \Return $m$
\EndProcedure
\\
\Procedure{updateAdaline}{$m$}
\State $m.w \gets m.w+ \eta (y - \langle m.w,x \rangle)x$
\State \Return $m$
\EndProcedure
\algstore{p2pegasos}
\end{algorithmic}
\end{minipage}
\hfil
\begin{minipage}{65mm}
\begin{algorithmic}[1]
\algrestore{p2pegasos}
\Procedure{initModel}{}
\State lastModel$.t \gets 0$
\State lastModel$.w \gets (0,\ldots,0)^T$
\State modelCache.add(lastModel)
\EndProcedure
\\
\Procedure{merge}{$m_1$,$m_2$}
\State $m.t \gets \max(m_1.t,m_2.t)$
\State $m.w \gets (m_1.w+m_2.w)/2$
\State \Return $m$
\EndProcedure
\end{algorithmic}
\end{minipage}
\end{algorithm}
It should be mentioned that---even though the number of ,,virtual'' models is growing
exponentially fast---the algorithm is not equivalent to bagging over an exponential
number of independent models.
In each gossip cycle, there are only $N$ independent updates occurring in the system overall
(where $N$ is the number of nodes),
and the effect of these updates is being aggregated rather efficiently.
In fact, as we will see in Section~\ref{sec:exp}, bagging over $N$ independent models 
actually outperforms the gossip learning algorithms.

\subsection{Pegasos}
\label{sec:ummu}

Here we discuss the adaptation of Pegasos (a linear SVM gradient method~\cite{pegasos10}
used for classification) into our gossip framework.
The components required for the adaptation are shown in Algorithm~\ref{alg:p2pegasos}, where
method {\sc updatePegasos} is simply taken from~\cite{pegasos10}.
For a complete implementation of the framework, one also needs to select an implementation of
{\sc createModel} from Algorithm~\ref{alg:createmodel}.
In the following, the three versions of a complete Pegasos-based implementation defined by these
options will be referred to as \pegasosrw, \pegasosmu, and \pegasosum.

The main difference between the Adaline perceptron and Pegasos is the context dependent update rule that is
different for correctly and incorrectly classified examples.
Due to this difference, there is no strict equivalence between averaging and voting, as in the case
of the previous section.
To see this, consider two models, $w_1$ and $w_2$, and an example $(x,y)$, and let $\bar w=(w_1+w_2)/2$.
In this case, updating $w_1$ and $w_2$ first, and then averaging them results in the same model as updating $\bar w$
if and only if both $w_1$ and $w_2$ classify $x$ in the same way (correctly or incorrectly).
This is because when updating $\bar w$, we virtually update both $w_1$ and $w_2$ in the same way, irrespective
of how they classify $x$ individually.

This seems to suggest that \pegasosum\ is a better choice.
We will test this hypothesis experimentally in Section~\ref{sec:exp}, where we will show that, surprisingly,
it is not always true.
The reason could be that \pegasosmu\ and \pegasosum\ are in fact very similar when we consider the
entire history of the distributed computation, as opposed to a single update step.
The histories of the models define a directed acyclic graph (DAG), where the nodes are merging operations,
and the edges correspond to the transfer of a model from one node to another.
In both cases, there is one update corresponding to each edge: the only difference is whether the
update occurs on the source node of the edge or on the target.
Apart from this, the edges of the DAG are the same for both methods.
Hence we see that \pegasosmu\ has the favorable property that the
updates that correspond to the incoming edges of a merge operation are done using independent
samples, while for \pegasosum\ they are performed with the same example.
Thus, \pegasosmu\ guarantees a greater independence of the models.

In the following we present our theoretical results for both \pegasosmu\ and \pegasosum.
We note that these results do not assume any coordination or synchronization; they are based
on a fully asynchronous communication model.
First let us formally define the optimization problem at hand, and let us
introduce some notation.

Let $S=\{(x_i,y_i):1\leq i\leq n, x_i\in\mathbb{R}^d, y_i\in\{+1,-1\}\}$ be a distributed training set
with one data point at each network node.
Let $f:\mathbb{R}^{d}\rightarrow\mathbb{R}$ be the objective function of the SVM learning problem 
(applying the L1 loss in the more general form proposed in Eq.~\eqref{eq:primal}):
\begin{equation}
\label{eq:svm}
\begin{split}
&f(w)=\min_{w}\frac{\lambda}{2}\|w\|^2+\frac{1}{n}\sum_{(x,y)\in\mathcal{S}}\ell(w;(x,y))\text{,}\\
&\text{where }\ell(w;(x,y)) = \max\{0,1-y\langle w,x\rangle\}
\end{split}
\end{equation}
Note that $f$ is strongly convex with a parameter $\lambda$~\cite{pegasos10}.
Let $w^{\star}$ denote the global optimum of $f$.
For a fixed data point $(x_i,y_i)$ we define
\begin{equation}
\label{eq:svmstoch}
f_i(w)=\frac{\lambda}{2}\|w\|^2+\ell(w;(x_i,y_i)),
\end{equation}
which is used to derive the update rule for the Pegasos algorithm.
Obviously, $f_i$ is $\lambda$ strongly convex as well, since it has the same form as
$f$ with $m=1$.

The update history of a model can be represented as a
binary tree, where the nodes are models, and the edges are defined by the direct ancestor relation.
Let us denote the direct ancestors of $w^{(i+1)}$ as $w_1^{(i)}$ and $w_2^{(i)}$.
These ancestors are averaged and then updated to obtain $w^{(i+1)}$ (assuming the MU variant).
Let the sequence $w^{(0)},\ldots,w^{(t)}$ be defined as the path in this history tree, for
which
\begin{equation}
\begin{split}
w^{(i)}=&\argmax_{w\in\{w_1^{(i)},w_2^{(i)}\}}\|w-w^{\star}\|,\\
i=&0,\ldots,t-1. 
\end{split}
\end{equation}
This sequence is well defined.
Let $(x_i,y_i)$ denote the training example, that was used
in the update step that resulted in $w^{(i)}$ in the series defined above.

\newtheorem{theorem}{Theorem}
\begin{theorem}[\pegasosmu\ convergence]\label{theroem:p2pegaosmuconv}
We assume that (1) each node receives an incoming message after any point in time within a finite time period (eventual update
assumption), (2) there is a subgradient $\nabla$ of the objective function such that $\|\nabla_w\|\leq G$ for every $w$.
Then,
\begin{equation}
\frac{1}{t}\sum_{i=1}^{t}f_i(\bar{w}^{(i)})-f_i(w^{\star})\leq \frac{G^2(log(t)+1)}{2\lambda t}
\end{equation}
where $\bar{w}^{(i)}=(w_1^{(i)}+w_2^{(i)})/2$.
\end{theorem}

\begin{proof}
During the running of the algorithm, let us pick any node on which at least one
subgradient update has been performed already.
There is such a node eventually, due to the eventual update assumption.
Let the model currently stored at this node be $w^{(t+1)}$. 

We know that $w^{(t+1)}=\bar{w}^{(t)}-\nabla^{(t)}/(\lambda t)$, where $\bar{w}^{(t)}=(w_1^{(t)}+w_2^{(t)})/2$ and 
where $\nabla^{(t)}$ is the subgradient of $f_t$.
From the $\lambda$-convexity of $f_t$ it follows that

\begin{multline}
\label{eq:lower}
f_t(\bar{w}^{(t)})-f_t(w^{\star})+\frac{\lambda}{2}\|\bar{w}^{(t)}-w^{\star}\|^2 \leq \\
\leq \langle \bar{w}^{(t)}-w^{\star},\nabla^{(t)}\rangle.
\end{multline}
On the other hand, the following inequality is also true, following from the definition of $\bar{w}^{(t+1)}$, $G$
and some algebraic rearrangements:

\begin{multline}
\label{eq:upper}
\langle \bar{w}^{(t)}-w^{\star},\nabla^{(t)}\rangle \leq \\
\leq \frac{\lambda t}{2}\|\bar w^{(t)}-w^{\star}\|^2 - \frac{\lambda t}{2}\|w^{(t+1)}-w^{\star}\|^2 + \frac{G^2}{2\lambda t}.
\end{multline}
Moreover, we can bound the distance of $\bar{w}^{(t)}$ from $w^{\star}$ with the distance of the
ancestor of $\bar{w}^{(t)}$ that is further away from $w^\star$ 
with the help of the Cauchy--Bunyakovsky--Schwarz inequality:

\begin{multline}
\label{eq:csb}
\|\bar w^{(t)}-w^{\star}\|^2 = \left\|\frac{w_1^{(t)}-w^{\star}}{2}+\frac{w_2^{(t)}-w^{\star}}{2}\right\|^2 \leq \\
\leq \|w^{(t)}-w^{\star}\|^2.
\end{multline}

From~\eqref{eq:lower}, ~\eqref{eq:upper}, ~\eqref{eq:csb} and the bound on the subgradients, 
we derive

\begin{multline}
\label{eq:bound_nonrec}
f_t(\bar{w}^{(t)})-f_t(w^{\star}) \leq\\
\leq \frac{\lambda (t-1)}{2}\|w^{(t)}-w^{\star}\|^2 - \frac{\lambda t}{2}\|w^{(t+1)}-w^{\star}\|^2 + \frac{G^2}{2\lambda t}.
\end{multline}

Note that this bound also holds for $w^{(i)}$, $1\leq i\leq t$. 
Summing up both sides of these $t$ inequalities, 
we get the following bound:

\begin{multline}
\label{eq:sum}
\sum_{i=1}^{t}f_i(\bar{w}^{(i)})-f_i(w^{\star}) \leq \\
\leq - \frac{\lambda t}{2}\|w^{(t+1)}-w^{\star}\|^2 + \frac{G^2}{2\lambda}\sum_{i=1}^{t}\frac{1}{i}\leq \frac{G^2(log(t)+1)}{2\lambda},
\end{multline}
from which the theorem follows after division by $t$.
\end{proof}

The bound in~\eqref{eq:sum} is analogous to the bound presented in~\cite{pegasos10}
in the analysis of the {\sc{Pegasos}}{} algorithm.
It basically means that the average error tends to zero.
To be able to show that the limit of the process is the optimum
of $f$, it is necessary that the samples involved in the series are
uniform random samples~\cite{pegasos10}.
Investigating the distribution of the samples is left to future work;
but we believe that the distribution closely approximates uniformity
for a large $t$, given the uniform random peer sampling that is applied.

%
For \pegasosum, an almost identical derivation leads us to a similar result (omitted
due to lack of space).

\section{Experimental Results}
\label{sec:exp}


We experiment with two algorithms: \pegasosum\ and \pegasosmu.
In addition, to shed light on the behavior of these algorithms, 
we include a number of baseline methods as well.
To perform the experiments, we used the {\sc PeerSim} event based P2P simulator~\cite{peersim}.

\subsection{Experimental Setup}

\paragraph{Baseline Algorithms.}
The first baseline we use is \pegasosrw.
If there is no message drop or message delay, then this is
equivalent to the Pegasos algorithm, since in cycle $t$ all peers will have
models that are the result of Pegasos learning on $t$ random examples.
In case of message delay and message drop failures, the number of samples
will be less than $t$, as a function of the drop probability and the delay.

We also examine two variants of {\em weighted bagging}.
The first variant ({\sc wb1}) is defined as
\begin{equation}
h_{WB1}(x,t)=\sgn(\sum_{i=1}^{N} \langle x,w_i^{(t)}\rangle),
\end{equation}
where $N$ is the number of nodes in the network, and the linear models $w_i^{(t)}$ are learned
with Pegasos over an independent sample of size $t$ of the training data.
This baseline algorithm can be thought of as the ideal utilization of
the $N$ independent updates performed in parallel by the $N$ nodes in the network in each cycle.
The gossip framework introduces dependencies among the models, so its performance
can be expected to be worse.

In addition, in the gossip framework a node has influence from only $2^t$ models on average
in cycle $t$.
To account for this handicap, we also use a second version of weighted bagging
({\sc wb2}):
\begin{equation}
h_{WB2}(x)=\sgn(\sum_{i=1}^{\min(2^t,N)} \langle x,w_i\rangle).
\end{equation}

The weighted bagging variants described above are not practical
alternatives, these algorithms serve as a baseline only.
The reason is that an actual implementation would require $N$ independent
models for prediction.
This could be achieved by \pegasosrw\ with a distributed prediction, which
would impose a large cost and delay for every
prediction.
This could also be achieved by all nodes running up to $O(N)$ instances of
\pegasosrw, and using the $O(N)$ local models for prediction; this is not feasible either.
In sum, the point that we want to make is that our gossip algorithm approximates
{\sc wb2} quite well using only a single message per node in each cycle, due to the
technique of merging models.

The last baseline algorithm we experiment with is {\sc perfect matching}.
In this algorithm we replace the peer sampling component of the gossip framework:
instead of all nodes picking random neighbors in each cycle, we create
a random perfect matching among the peers so that every peer receives exactly one message.
Our hypothesis was that---since this variant increases the efficiency of mixing---it
will maintain a higher diversity of models, and so a better performance can be
expected due to the ``virtual bagging'' effect we explained previously.
Note that this algorithm is not intended to be practical either.

\paragraph{Data Sets.}

\begin{table*}
\caption{The main properties of the data sets, and the prediction error (0-1 error) of the
baseline sequential algorithm. In the case of Malicious URLs dataset the results of the full feature set are shown in parentheses.}
\centering
\label{dbprops}
\begin{center}
\begin{tabular}{|l||c|c|c|c|}
\hline
 & Reuters & SpamBase & Malicious URLs (10)\\
\hline\hline
Training set size & 2,000 & 4,140 & 2,155,622 \\
Test set size & 600 & 461 & 240,508 \\
Number of features & 9,947 & 57 & 10 \\
Class label ratio & 1,300:1,300 & 1,813:2,788 & 792,145:1,603,985 \\
\hline
Pegasos 20,000 iter. & 0.025 & 0.111 & 0.080 (0.081) \\
\hline
\end{tabular}
\end{center}
\end{table*}

We used three different data sets: Reuters~\cite{reuters}, Spambase, and the Malicious URLs~\cite{ma-icml09} data sets,
which were obtained from the UCI database repository~\cite{ucirepo}.
These data sets are of different types including small and large sets containing a small or large number of features.
Table~\ref{dbprops} shows the main properties of these data sets, as well as the prediction performance of the Pegasos algorithm.

The original Malicious URLs data set has a huge number of features ($\sim$ 3,000,000), therefore we first performed a feature reduction step so that we can carry out simulations. Note that the message size in our algorithm depends on the number of features, therefore in a real application this step might also be useful in such extreme cases. We applied the well-known  correlation coefficient method for each feature with the class label, and kept the ten features with the maximal absolute values. If necessary, this calculation can also be carried out in a gossip-based fashion~\cite{JMB05}, but we performed it offline. The effect of this dramatic reduction on the prediction performance is shown in Table~\ref{dbprops}, where Pegasos results on the full feature set are shown in parenthesis.

\paragraph{Using the local models for prediction.}

An important aspect of our protocol is that every node has at least one model available
locally, and thus all the nodes can perform a prediction.
Moreover, since the nodes can remember the models that pass through them at no communication cost,
we cheaply implement a simple voting mechanism, where nodes will use more than one
model to make predictions.
Algorithm~\ref{alg:pred} shows the procedures used for prediction in the original case, and in the case
of voting.
\begin{algorithm}[bt]
\caption{Local prediction procedures}
\label{alg:pred}
\begin{minipage}{6cm}
\begin{algorithmic}[1]
\Procedure{predict}{$x$}
\State $w \gets$ modelCache.freshest()
\State \textbf{return} sign($\left\langle w, x \right\rangle$)
\EndProcedure
\algstore{pred}
\end{algorithmic}
\end{minipage}
\hfil
\begin{minipage}{8cm}
\begin{algorithmic}[1]
\algrestore{pred}
\Procedure{votedPredict}{$x$}
\State pRatio $\gets 0$
\For{$m \in$ modelCache}
\If{sign$(\left\langle m.w, x \right\rangle) \ge 0$}
\State pRatio $\gets$ pRatio $+ 1$
\EndIf
\EndFor
\State \textbf{return} sign(pRatio/modelCache.size()$- 0.5$)
\EndProcedure
\end{algorithmic}
\end{minipage}
\end{algorithm}
Here the vector $x$ is the unseen example to be classified.
In the case of linear models, the classification is simply the sign of the inner product
with the model, which essentially describes on which side of the hyperplane the given point lies.
In our experiments we used a cache of size 10.

\paragraph{Evaluation metric.}


The evaluation metric we focus on is prediction error. To measure prediction error, we need to split the datasets into training sets and test sets. The proportions of this splitting are shown in Table~\ref{dbprops}.
In our experiments with \pegasosmu\ and \pegasosum\ we track the misclassification ratio over the test set of
100 randomly selected peers.
The misclassification ratio of a model is simply the number of the misclassified test examples divided by the number of all test examples, which is the so called 0-1 error. 

For the baseline algorithms we used all the available models for calculating the error rate,
which equals the number of training samples.
From the Malicious URLs database we used only 10,000 examples selected at random,
to make the evaluation computationally feasible.
Note, that we found that increasing the number of examples beyond 10,000 does nor result in a noticeable
difference in the observed behavior.

We also calculated the similarities between the models circulating in the network,
using the cosine similarity measure.
We calculated the similarity between all pairs of models,
and calculated the average.
This metric is useful to study the speed at which the actual models converge.
Note that under uniform sampling it is known that all models converge to an optimal model.

\paragraph{Modeling failure.}
In a set of experiments we model extreme message drop and message delay.
Drop probability is set to be $0.5$.
This can be considered an extremely large drop rate.
Message delay is modeled as a uniform random delay from the interval [$\Delta$, $10\Delta$], where $\Delta$ is the gossip period in Algorithm~\ref{alg:gossiplearn}.
This is also an extreme delay, orders of magnitudes higher than what can be expected in a realistic scenario, except if
$\Delta$ is very small.
We also model realistic churn based on probabilistic models in~\cite{p2pchurn}. Accordingly,
we approximate online session length with a lognormal distribution, and we approximate the parameters of the distribution using a maximum likelihood estimate based
on a trace from a private BitTorrent community called FileList.org obtained from Delft
University of Technology~\cite{roozenburg-msc}.
We set the offline session lengths so that at any moment
in time 90\% of the peers are online. In addition, we assume that when a peer comes
back online, it retains its state that it had at the time of leaving the network.

\subsection{Results and Discussion}

\begin{figure*}[tb]
\begin{center}
\includegraphics[width=.33\textwidth]{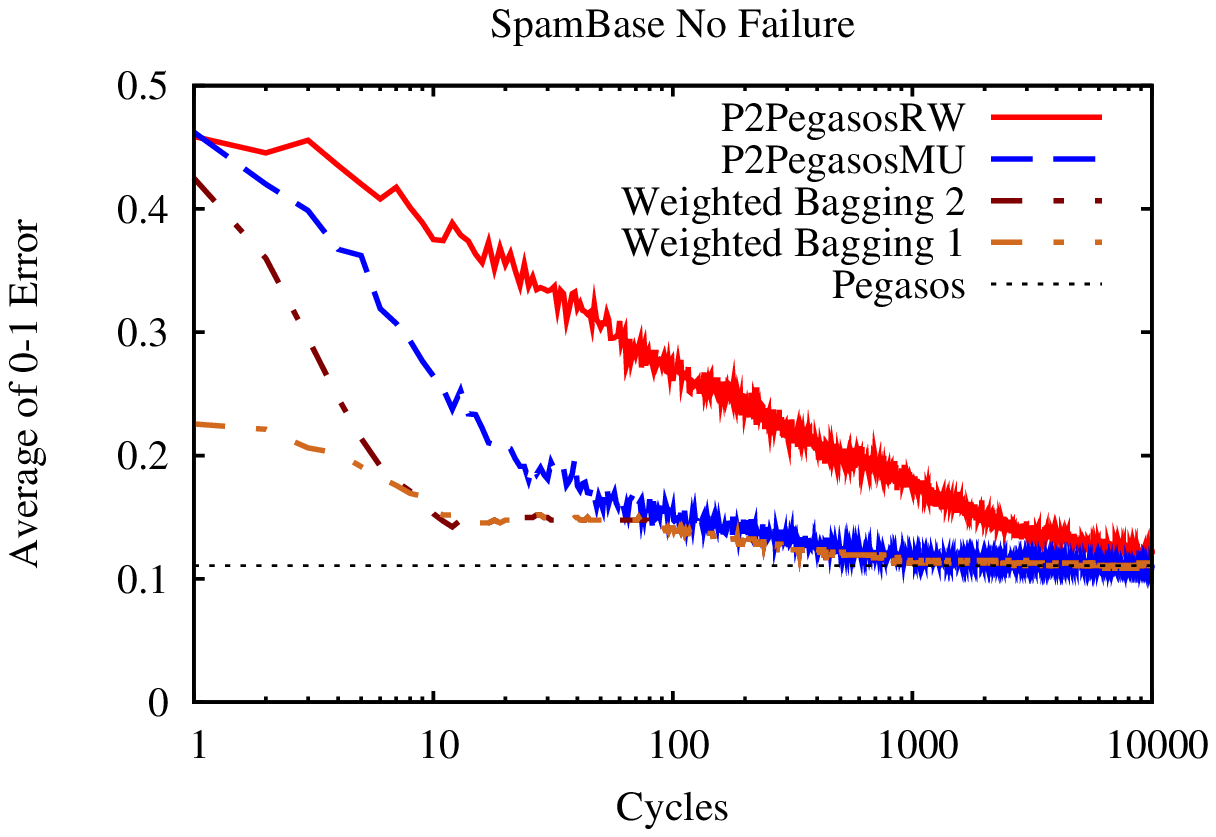}~%
\includegraphics[width=.33\textwidth]{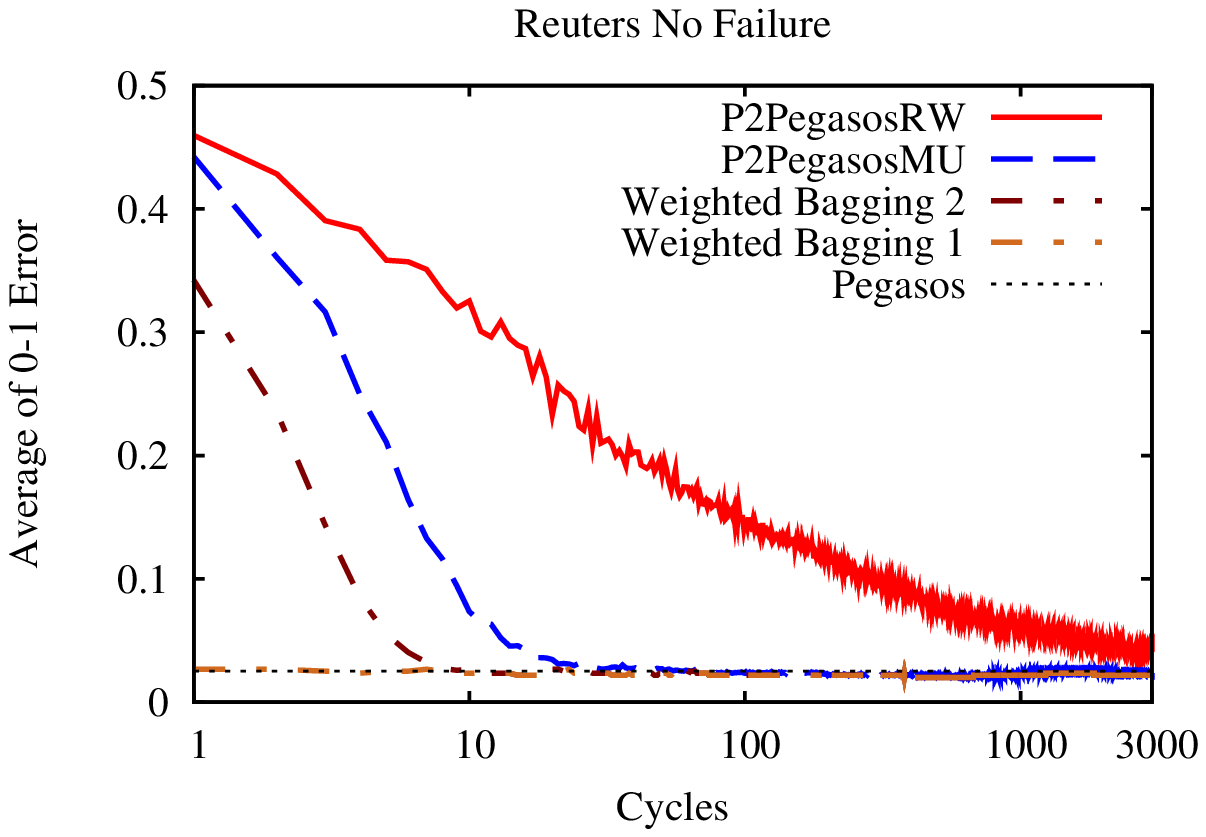}~%
\includegraphics[width=.33\textwidth]{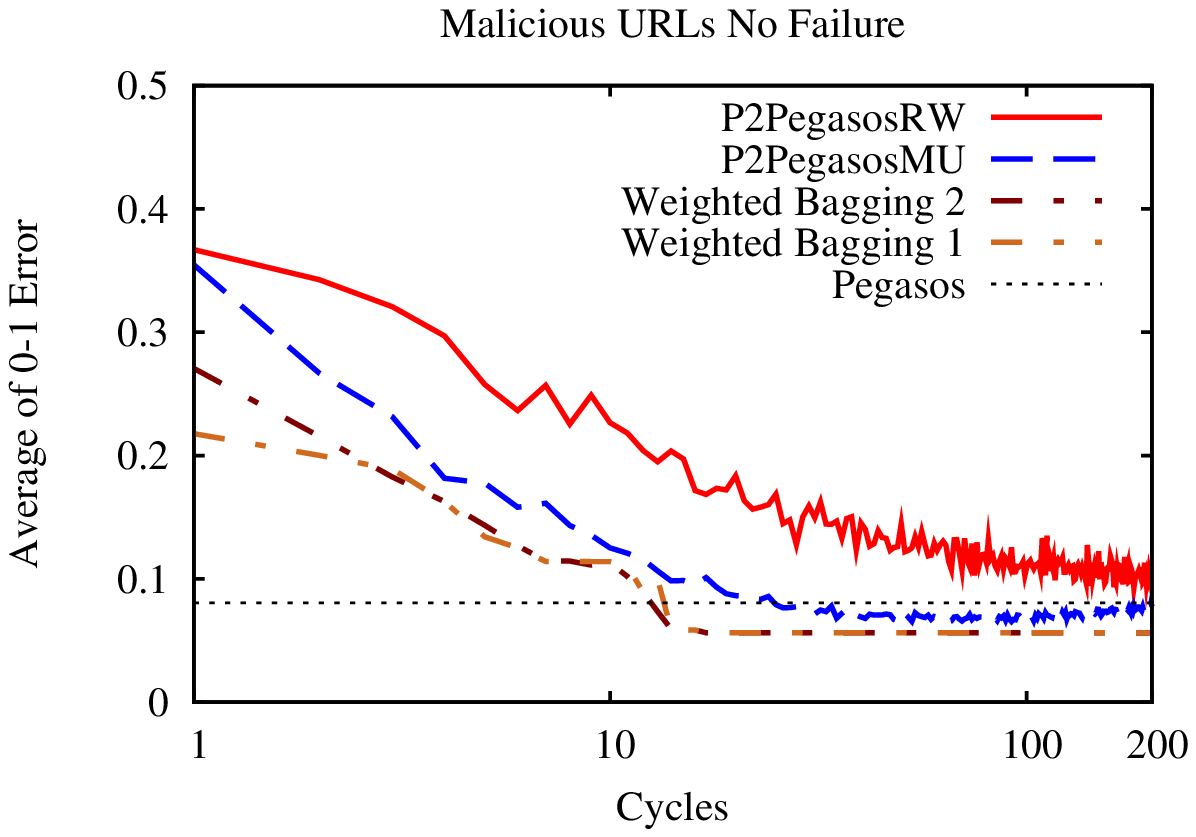}
\includegraphics[width=.33\textwidth]{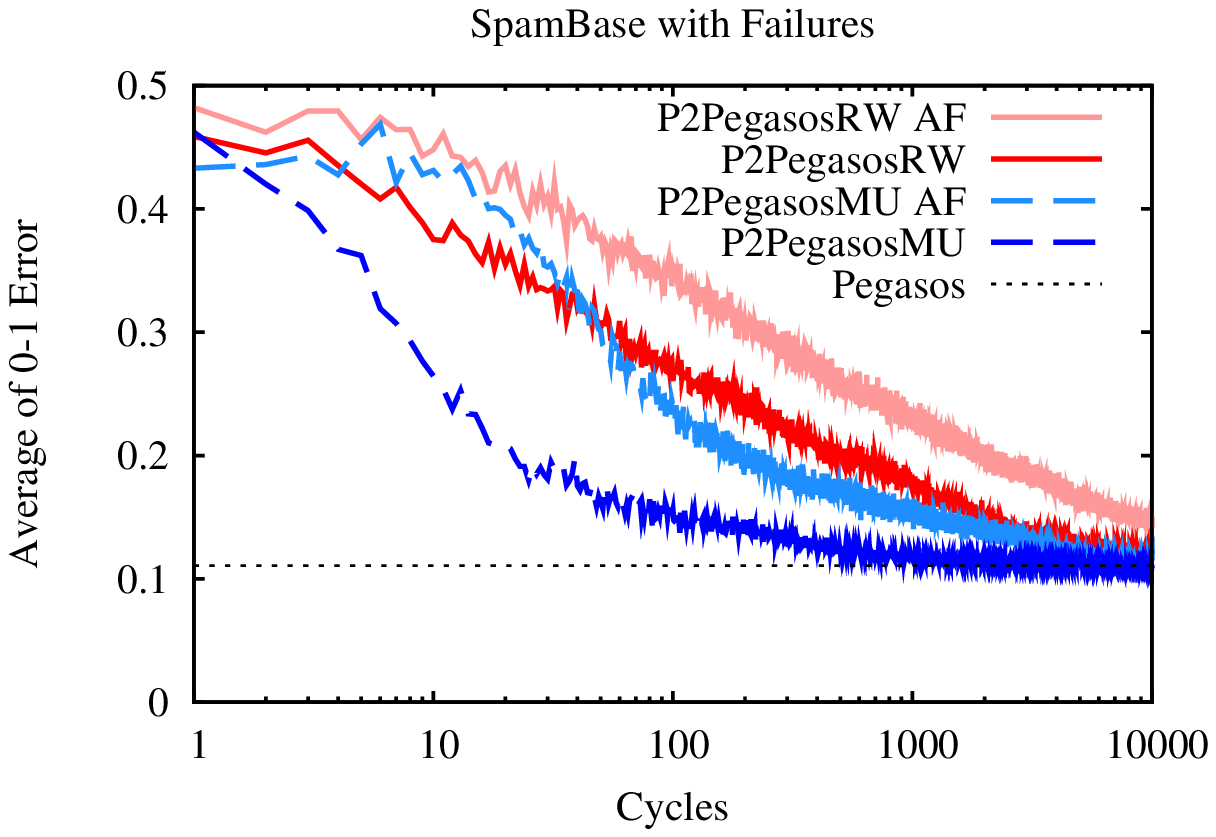}~%
\includegraphics[width=.33\textwidth]{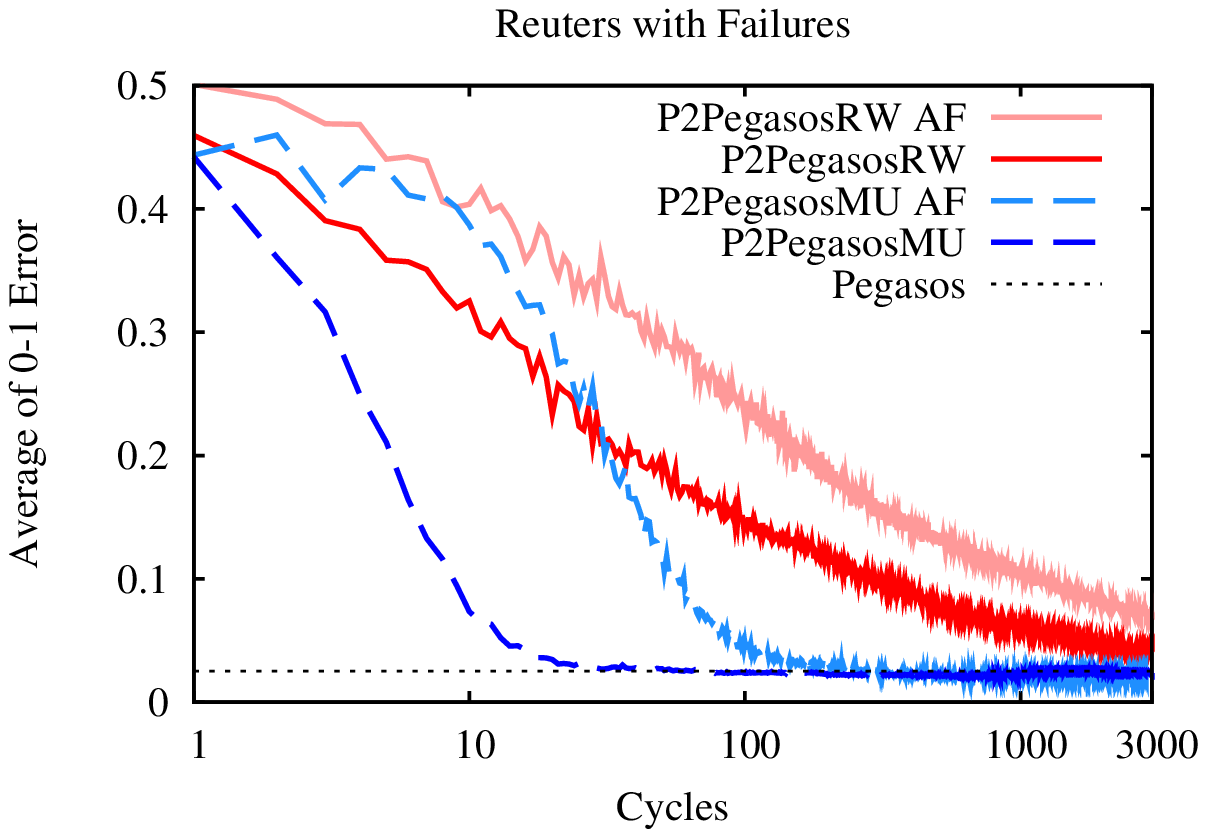}~%
\includegraphics[width=.33\textwidth]{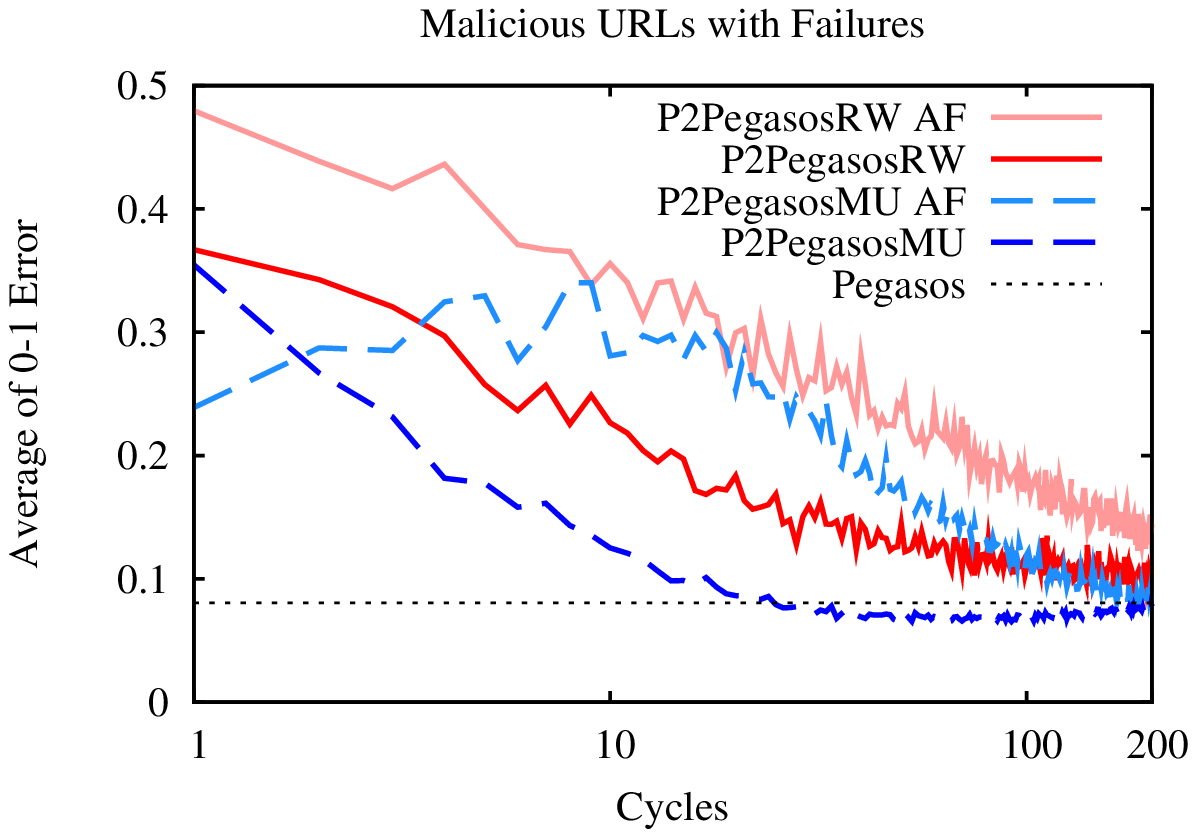}
\end{center}
\caption{Experimental results without failure (upper row) and with extreme failure (lower row). AF means
all possible failures are modeled.}
\label{fig:1}
\end{figure*}

\begin{figure*}[tb]
\includegraphics[width=.33\textwidth]{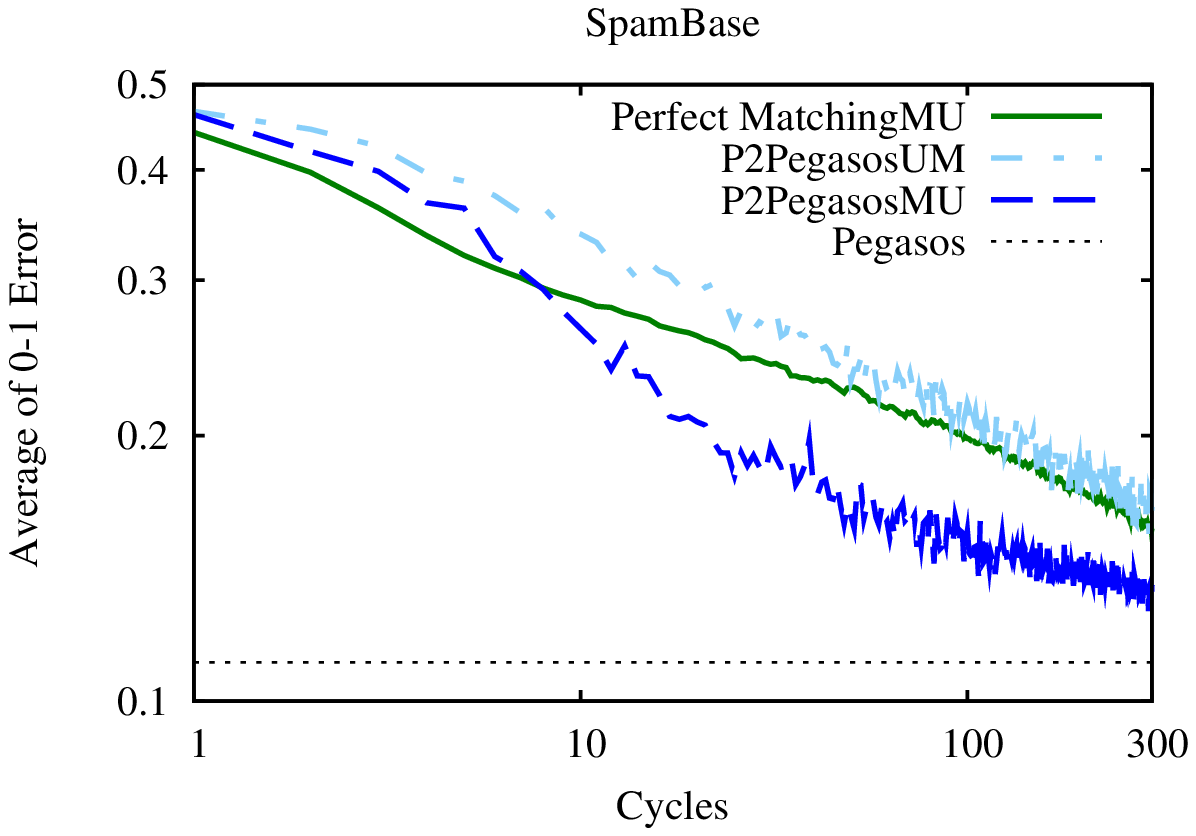}~%
\includegraphics[width=.33\textwidth]{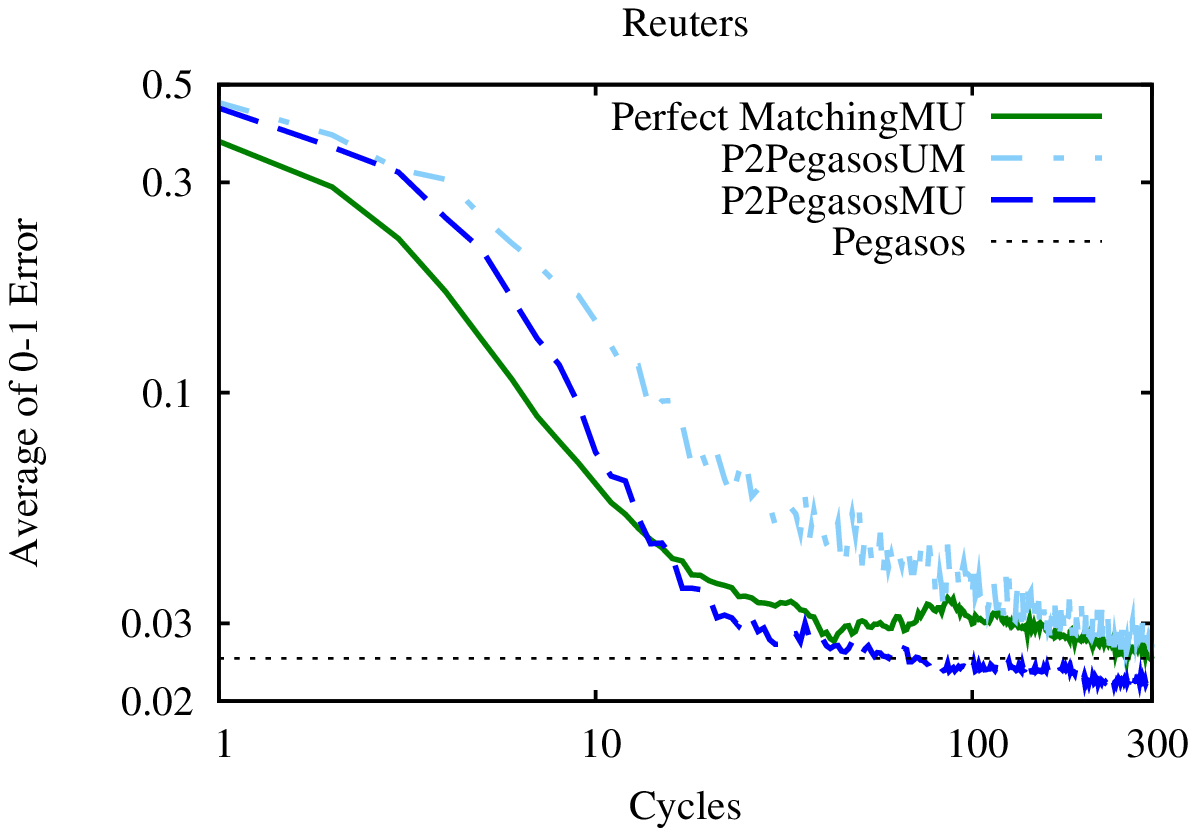}~%
\includegraphics[width=.33\textwidth]{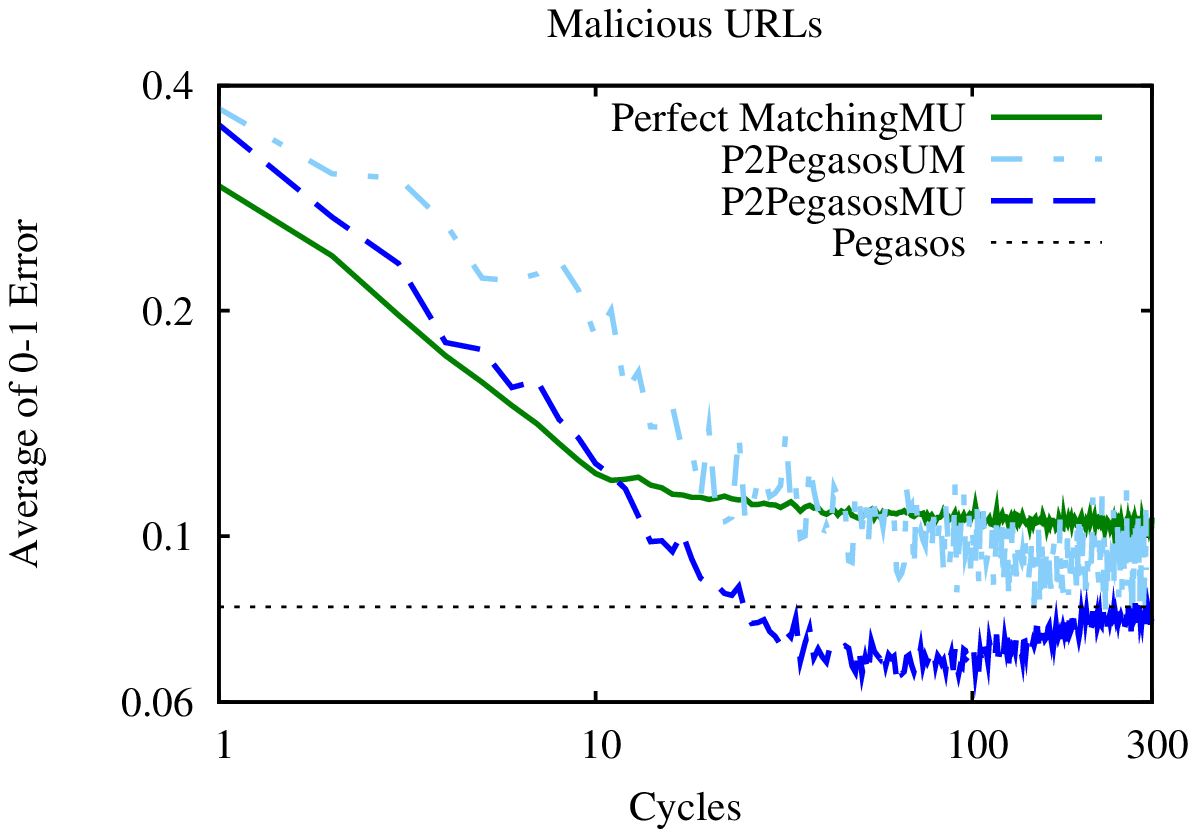}
\includegraphics[width=.33\textwidth]{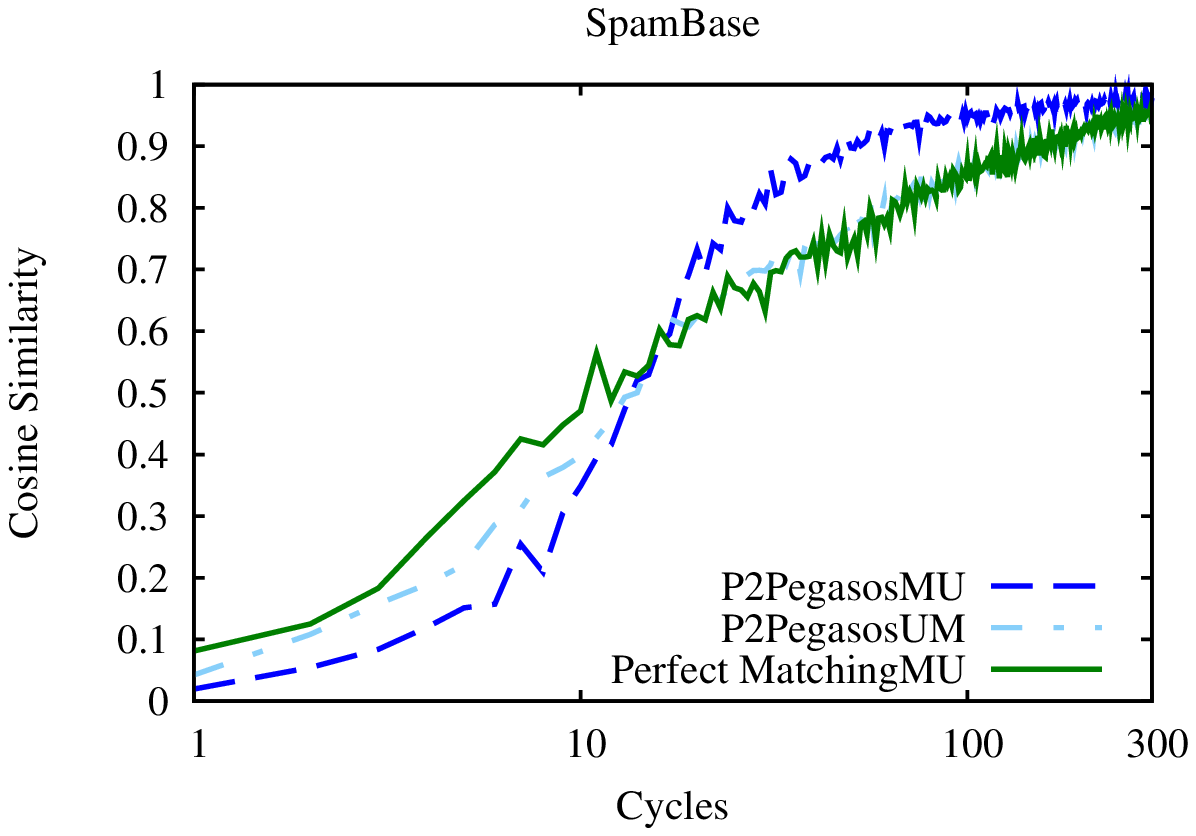}~%
\includegraphics[width=.33\textwidth]{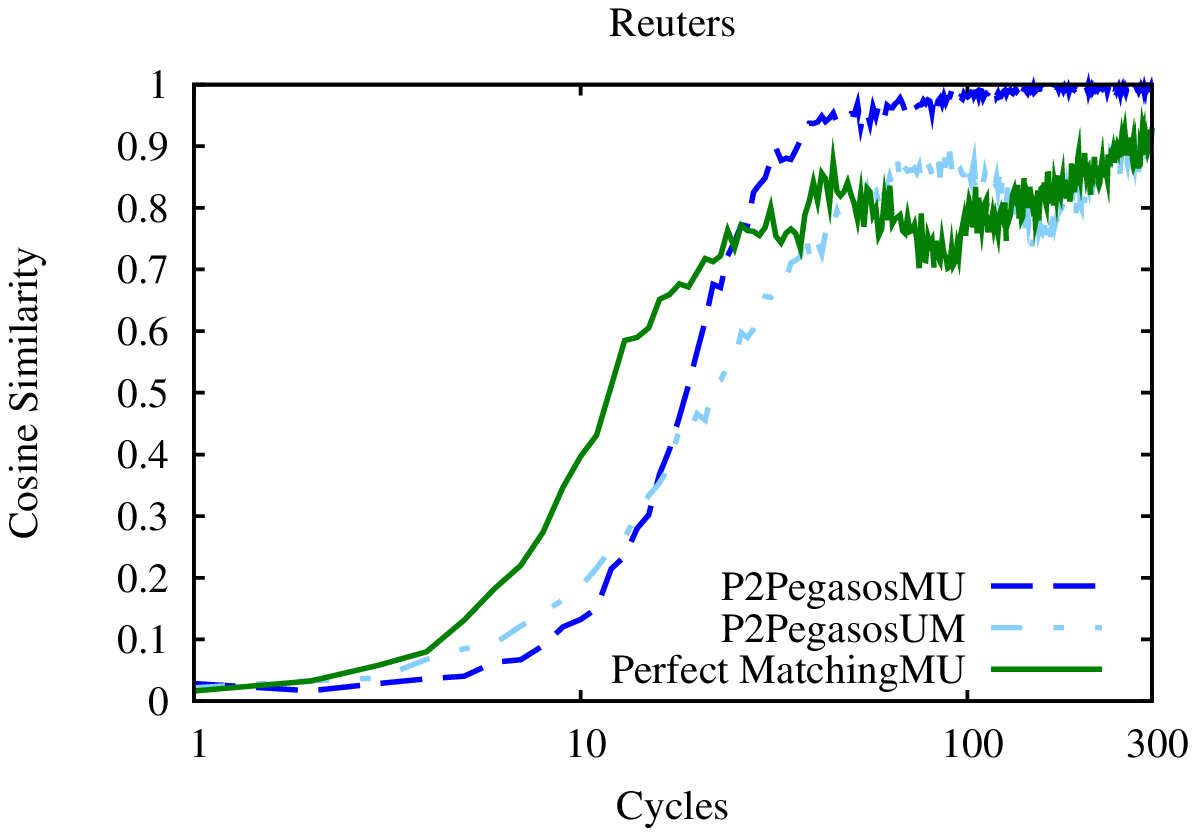}~%
\includegraphics[width=.33\textwidth]{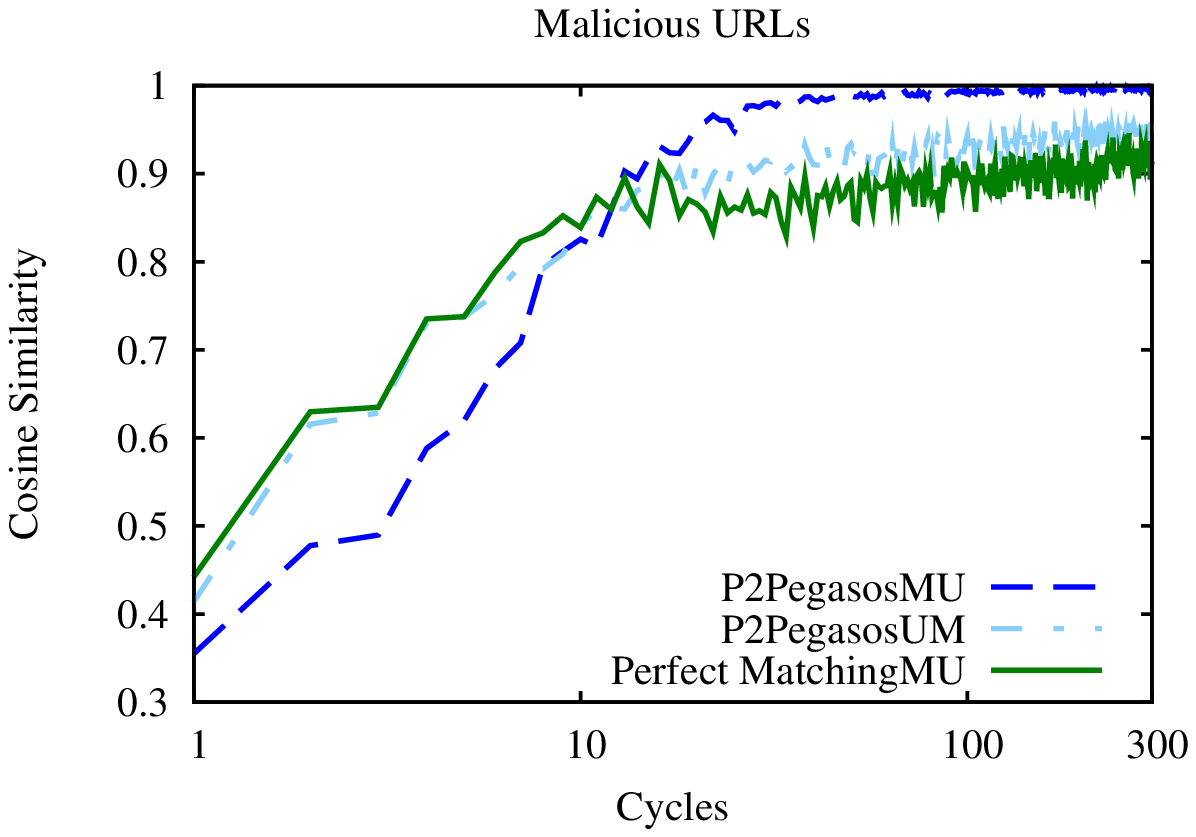}
\caption{Prediction error (upper row) and model similarity (lower row) with {\sc perfect matching} and \pegasosum.}
\label{fig:2}
\end{figure*}

The experimental results for prediction without local voting are shown in Figures~\ref{fig:1} and~\ref{fig:2}.
Note that all variants can be mathematically proven to converge to the same result,
so the difference is in convergence speed only.
Bagging can temporarily outperform a single instance of
Pegasos, but after enough training samples, all models become almost identical, so
the advantage of voting disappears.

In Figure~\ref{fig:1} we can see that our hypothesis about the relationship of the
performance of the gossip algorithms and the baselines is validated:
the standalone Pegasos algorithm is the slowest, while the two variants of weighted bagging are the
fastest.
\pegasosmu\ approximates {\sc wb2} quite well, with some delay, so we can use {\sc wb2} as a heuristic model
of the behavior of the algorithm.
Note that the convergence is several orders of magnitude faster than that of Pegasos (the
plots have a logarithmic scale).

Figure~\ref{fig:1} also contains results from our extreme failure scenario.
We can observe that the difference in convergence speed is mostly accounted for by the
increased message delay.
The effect of the delay is that all messages wait 5 cycles on average before being delivered,
so the convergence is proportionally slower. In addition, half of the messages get lost too,
which adds another factor of about 2 to the convergence speed.
Apart from slowing down, the algorithms still converge to the correct value despite
the extremely unreliable environment, as was expected.

Figure~\ref{fig:2} illustrates the difference between the UM and MU variants.
Here we model no failures.
In Section~\ref{sec:ummu} we pointed out that---although the UM version
looks favorable when considering a single node---when looking at the
full history of the learning process \pegasosmu\ maintains more independence between
the models.
Indeed, the MU version clearly performs better according to our experiments.
We can also observe that the UM version shows a lower level of model similarity
in the system, which probably has to do with the slower convergence.

\begin{sloppypar}
In Figure~\ref{fig:2} we can see the performance of the perfect matching variant of \pegasosmu\ as well.
Contrary to our expectations, perfect
matching does not clearly improve performance, apart from the first few cycles.
It is also interesting to observe, that model similarity is correlated
to prediction performance also in this case.
We also note, that in the case of the Adaline-based gossip learning implementation
perfect matching is clearly better than random peer sampling (not shown).
This means that this behavior is due to
the context-dependence of the update rule discussed in~\ref{sec:ummu}.
\end{sloppypar}

\begin{figure*}[tb]
\begin{center}
\includegraphics[width=.33\textwidth]{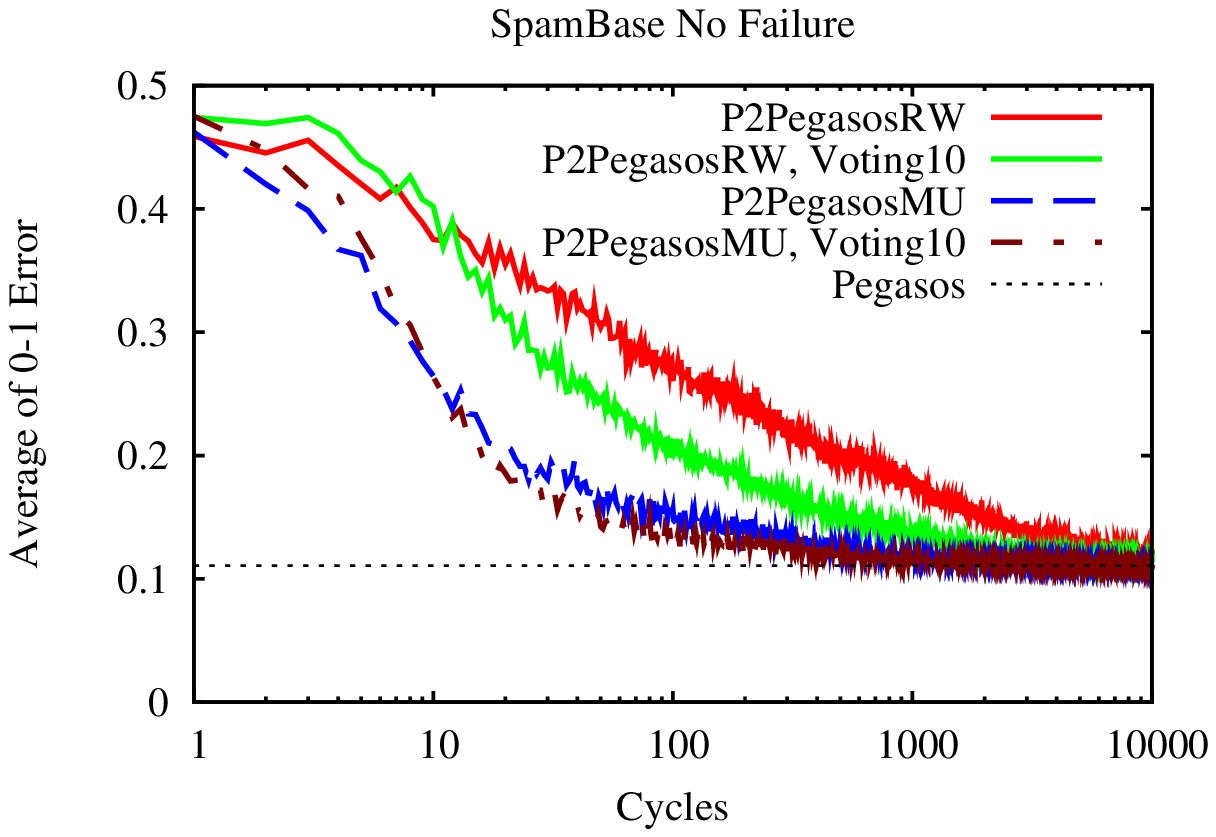}~%
\includegraphics[width=.33\textwidth]{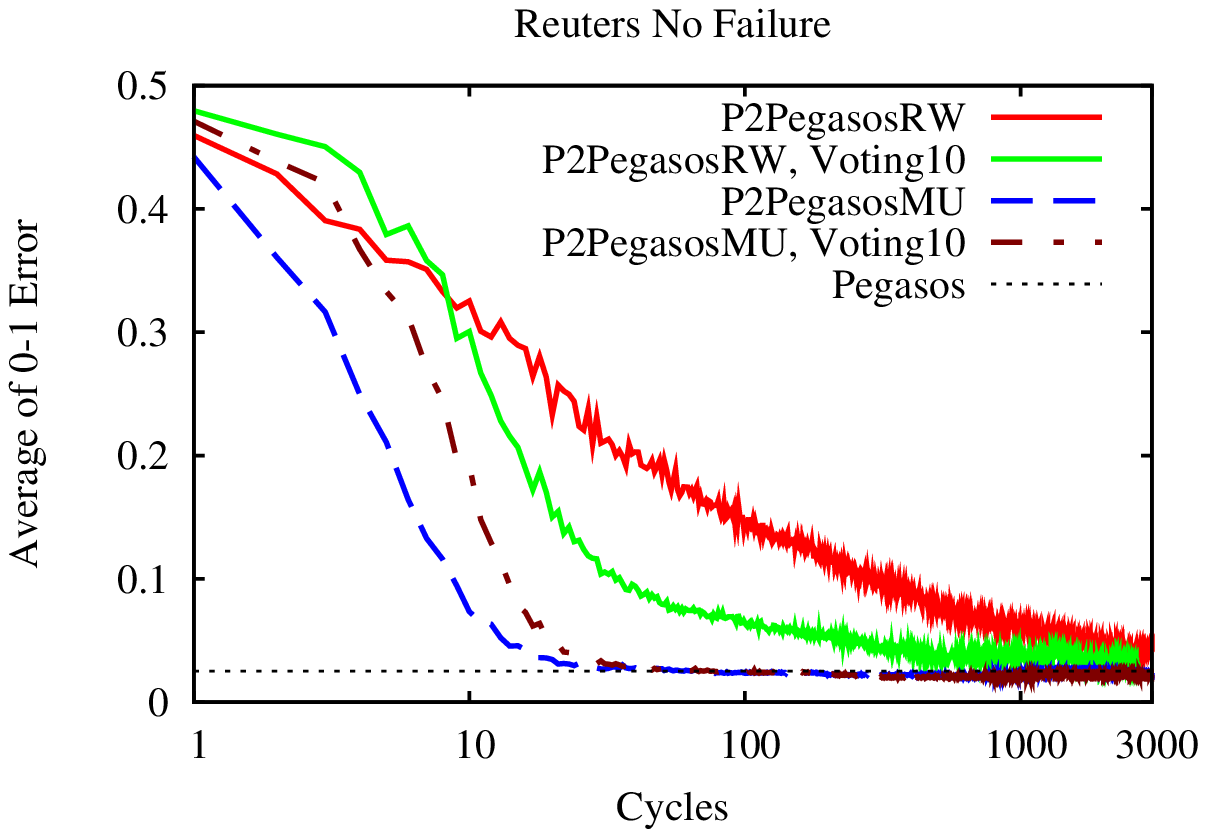}~%
\includegraphics[width=.33\textwidth]{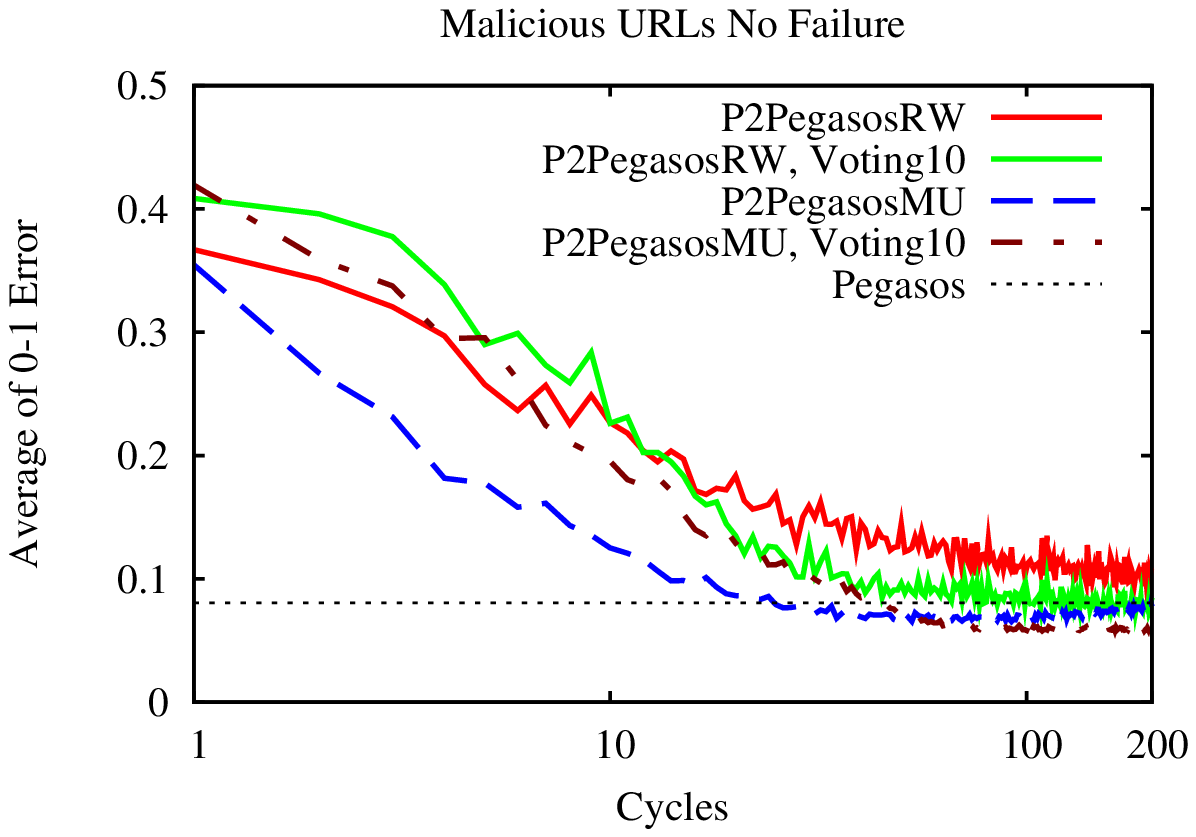}
\includegraphics[width=.33\textwidth]{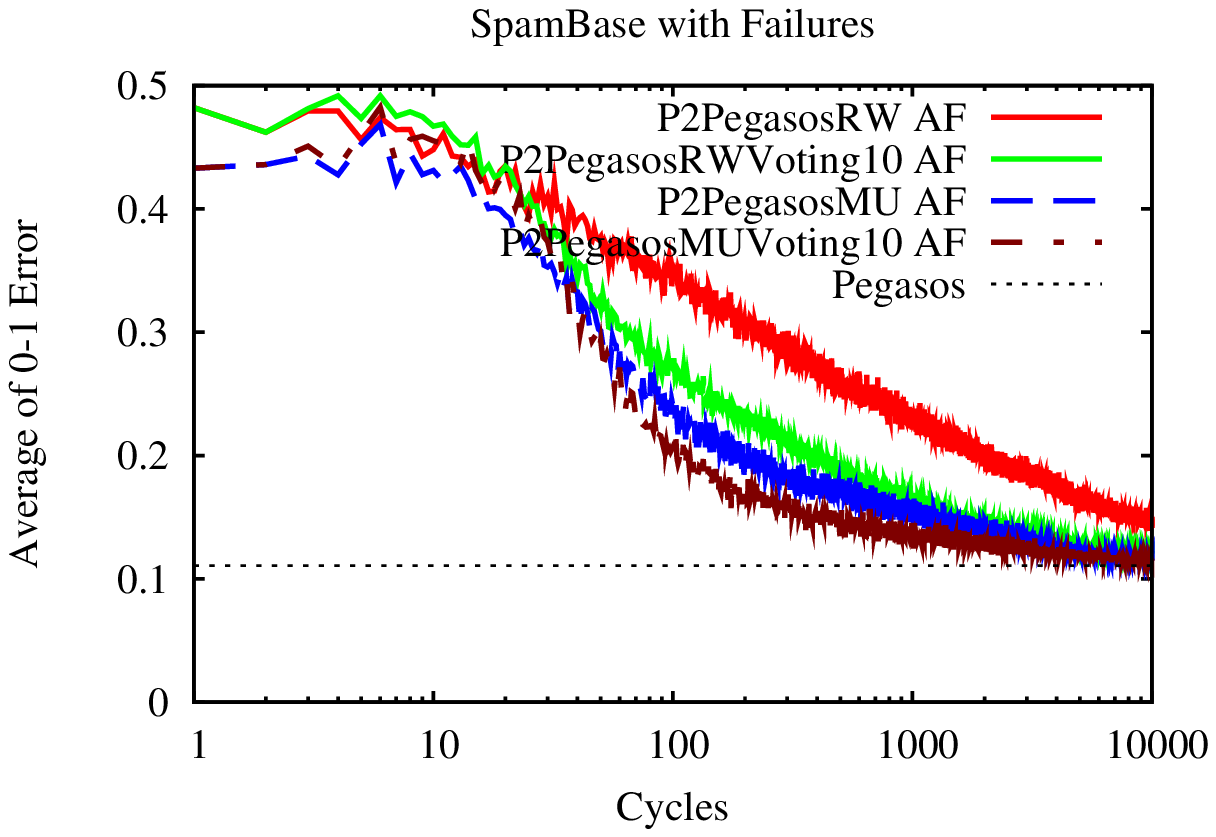}~%
\includegraphics[width=.33\textwidth]{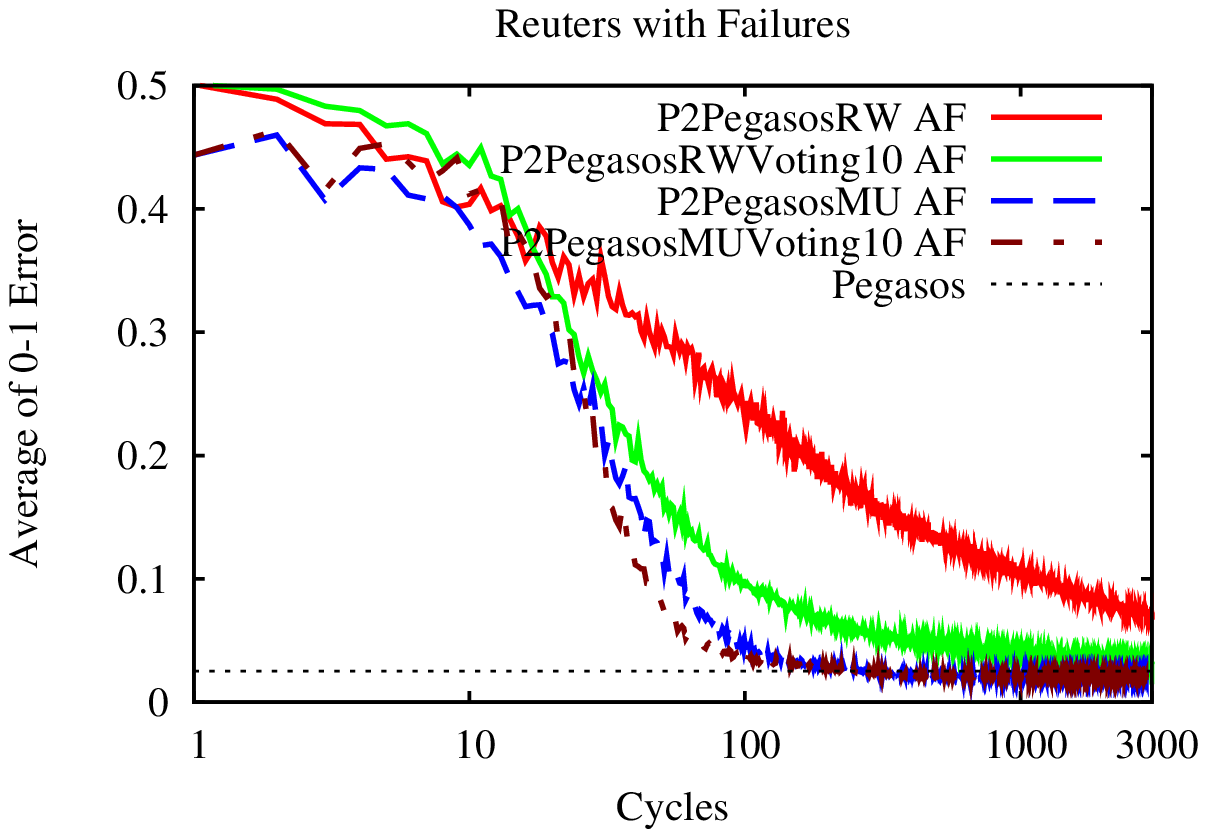}~%
\includegraphics[width=.33\textwidth]{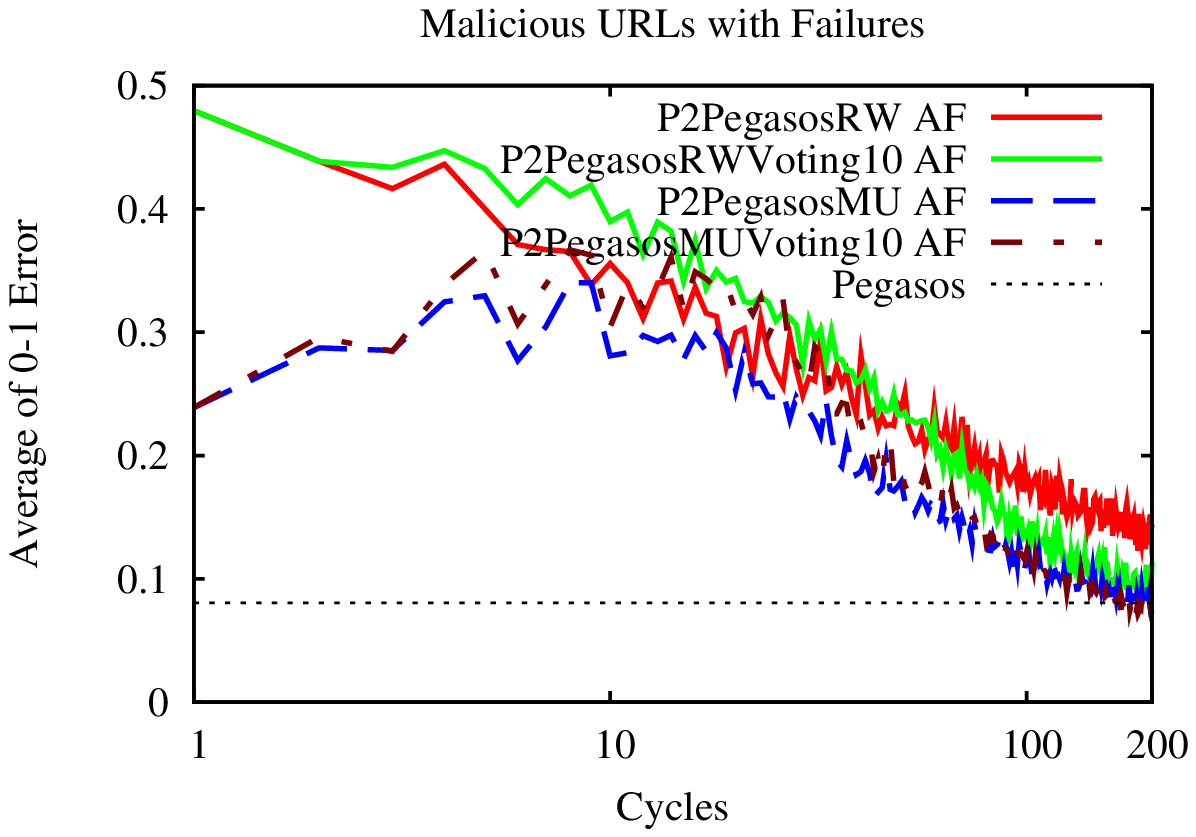}
\end{center}
\caption{Experimental results applying local voting without failure (upper row) and with extreme failure (lower row).}
\label{fig:3}
\end{figure*}

The results with local voting are shown in Figure~\ref{fig:3}.
The main conclusion is that voting results in a significant improvement when applied along with {\sc P2PegasosRW},
the learning algorithm that does not apply merging.
When merging is applied, the improvement is less dramatic.
In the first few cycles, voting can result in a slight degradation of performance.
This could be expected, since the models in the local caches are trained on fewer samples on average than the freshest
model in the cache.
Overall, since voting is for free, it is advisable to use it.

\section{Conclusions}
\label{sec:concl}


We proposed gossip learning as a generic approach to learn models of fully
distributed data in large scale P2P systems.
The basic idea of gossip learning is that many models perform a random walk
over the network, while being updated at every node they visit, and while
being combined (merged) with other models they encounter.
We presented an instantiation of gossip learning based on the Pegasos
algorithm.
The algorithm was shown to be extremely robust to message drop and message
delay, furthermore, a very significant speedup was demonstrated w.r.t.
the baseline Pegasos algorithm due to the model merging technique and the prediction
algorithm that is based on local voting.

The algorithm makes it possible to compute predictions locally at
every node in the network at any point in time, yet the message complexity is acceptable:
every node sends one model in each gossip cycle.
The main features that differentiate this approach from related work
are the focus on fully distributed data and its modularity, generality, 
and simplicity.

An important promise of the approach is the support for privacy preservation,
since data samples are not observed directly.
Although in this paper we did not focus on this aspect,
it is easy to see that the only feasible attack is 
the multiple forgery attack~\cite{multiple-forgery-attack}, where the local
sample is guessed based on sending specially crafted models to nodes and observing
the result of the update step.
This is very hard to do even without any extra measures, given that models perform
random walks based on local decisions, and that merge operations are performed as well. 
This short informal reasoning motivates our
ongoing work towards understanding and enhancing the privacy-preserving properties of gossip learning.



\bibliography{references}

\begin{thebibliography}{10}
\providecommand{\url}[1]{#1}
\csname url@samestyle\endcsname
\providecommand{\newblock}{\relax}
\providecommand{\bibinfo}[2]{#2}
\providecommand{\BIBentrySTDinterwordspacing}{\spaceskip=0pt\relax}
\providecommand{\BIBentryALTinterwordstretchfactor}{4}
\providecommand{\BIBentryALTinterwordspacing}{\spaceskip=\fontdimen2\font plus
\BIBentryALTinterwordstretchfactor\fontdimen3\font minus
  \fontdimen4\font\relax}
\providecommand{\BIBforeignlanguage}[2]{{%
\expandafter\ifx\csname l@#1\endcsname\relax
\typeout{** WARNING: IEEEtran.bst: No hyphenation pattern has been}%
\typeout{** loaded for the language `#1'. Using the pattern for}%
\typeout{** the default language instead.}%
\else
\language=\csname l@#1\endcsname
\fi
#2}}
\providecommand{\BIBdecl}{\relax}
\BIBdecl

\bibitem{tribler}
J.~A. Pouwelse, P.~Garbacki, J.~Wang, A.~Bakker, J.~Yang, A.~Iosup, D.~H.~J.
  Epema, M.~Reinders, M.~R. van Steen, and H.~J. Sips, ``{TRIBLER:} a
  social-based peer-to-peer system,'' \emph{Concurrency and Computation:
  Practice and Experience}, vol.~20, no.~2, pp. 127--138, 2008.

\bibitem{gossple-ebdt10}
X.~Bai, M.~Bertier, R.~Guerraoui, A.-M. Kermarrec, and V.~Leroy, ``Gossiping
  personalized queries,'' in \emph{Proceedings of the 13th International
  Conference on Extending Database Technology ({EBDT'10})}, 2010.

\bibitem{peerson}
S.~Buchegger, D.~Schi\"{o}berg, L.-H. Vu, and A.~Datta, ``{PeerSoN}: {P2P}
  social networking: early experiences and insights,'' in \emph{Proceedings of
  the Second ACM EuroSys Workshop on Social Network Systems ({SNS'09})}.\hskip
  1em plus 0.5em minus 0.4em\relax New York, NY, USA: ACM, 2009, pp. 46--52.

\bibitem{p2pwormdetection}
S.~G. Cheetancheri, J.~M. Agosta, D.~H. Dash, K.~N. Levitt, J.~Rowe, and E.~M.
  Schooler, ``A distributed host-based worm detection system,'' in
  \emph{Proceedings of the 2006 {SIGCOMM} workshop on Large-scale attack
  defense ({LSAD'06})}.\hskip 1em plus 0.5em minus 0.4em\relax New York, NY,
  USA: ACM, 2006, pp. 107--113.

\bibitem{ormandi2011asynchronP2PDM}
R.~Orm{\'a}ndi, I.~Heged\H{u}s, and M.~Jelasity, ``Asynchronous peer-to-peer
  data mining with stochastic gradient descent,'' in \emph{17th International
  European Conference on Parallel and Distributed Computing (Euro-Par 2011)},
  ser. Lecture Notes in Computer Science, vol. 6852.\hskip 1em plus 0.5em minus
  0.4em\relax Springer-Verlag, 2011, pp. 528--540.

\bibitem{astrolabe}
R.~van Renesse, K.~P. Birman, and W.~Vogels, ``Astrolabe: A robust and scalable
  technology for distributed system monitoring, management, and data mining,''
  \emph{ACM Transactions on Computer Systems}, vol.~21, no.~2, pp. 164--206,
  May 2003.

\bibitem{JMB05}
M.~Jelasity, A.~Montresor, and O.~Babaoglu, ``Gossip-based aggregation in large
  dynamic networks,'' \emph{{ACM} Transactions on Computer Systems}, vol.~23,
  no.~3, pp. 219--252, August 2005.

\bibitem{boyd05}
S.~Boyd, A.~Ghosh, B.~Prabhakar, and D.~Shah, ``Randomized gossip algorithms,''
  \emph{IEEE Transactions on Information Theory}, vol.~52, no.~6, pp.
  2508--2530, 2006.

\bibitem{pentland12}
A.~S. Pentland, ``Society's nervous system: Building effective government,
  energy, and public health systems,'' \emph{Computer}, vol.~45, no.~1, pp.
  31--38, Jan. 2012.

\bibitem{mobiscope07}
T.~Abdelzaher, Y.~Anokwa, P.~Boda, J.~Burke, D.~Estrin, L.~Guibas, A.~Kansal,
  S.~Madden, and J.~Reich, ``Mobiscopes for human spaces,'' \emph{Pervasive
  Computing, IEEE}, vol.~6, no.~2, pp. 20--29, april-june 2007.

\bibitem{mobilesensing10}
N.~Lane, E.~Miluzzo, H.~Lu, D.~Peebles, T.~Choudhury, and A.~Campbell, ``A
  survey of mobile phone sensing,'' \emph{Communications Magazine, IEEE},
  vol.~48, no.~9, pp. 140--150, Sep. 2010.

\bibitem{diaspora}
Diaspora, ``{https://joindiaspora.com/}.''

\bibitem{ma-icml09}
J.~Ma, L.~K. Saul, S.~Savage, and G.~M. Voelker, ``Identifying suspicious
  {URL}s: an application of large-scale online learning,'' in \emph{Proceedings
  of the 26th Annual International Conference on Machine Learning ({ICML
  '09})}.\hskip 1em plus 0.5em minus 0.4em\relax New York, NY, USA: ACM, 2009,
  pp. 681--688.

\bibitem{bottou-nips07tut}
L.~Bottou, ``The tradeoffs of large-scale learning,'' 2007, tutorial at the
  21st Annual Conference on Neural Information Processing Systems (NIPS),
  \url{http://leon.bottou.org/talks/largescale}.

\bibitem{bottou-lecun-2004}
L.~Bottou and Y.~{LeCun}, ``Large scale online learning,'' in \emph{Advances in
  Neural Information Processing Systems 16}, S.~Thrun, L.~Saul, and
  B.~{Sch\"{o}lkopf}, Eds.\hskip 1em plus 0.5em minus 0.4em\relax Cambridge,
  MA: MIT Press, 2004.

\bibitem{duda-hart-stork-book}
R.~O. Duda, P.~E. Hart, and D.~G. Stork, \emph{Pattern Classification},
  2nd~ed.\hskip 1em plus 0.5em minus 0.4em\relax Wiley-Interscience, 2000.

\bibitem{svm-book}
N.~Cristianini and J.~Shawe-Taylor, \emph{An introduction to Support Vector
  Machines and other kernel-based learning methods}.\hskip 1em plus 0.5em minus
  0.4em\relax Cambridge University Press, 2000.

\bibitem{pegasos10}
S.~Shalev-Shwartz, Y.~Singer, N.~Srebro, and A.~Cotter, ``Pegasos: primal
  estimated sub-gradient solver for {SVM},'' \emph{Mathematical Programming B},
  2010.

\bibitem{primal-svm07}
O.~Chapelle, ``Training a support vector machine in the primal,'' \emph{Neural
  Computation}, vol.~19, pp. 1155--1178, May 2007.

\bibitem{ensemble-learning10}
L.~Rokach, ``Ensemble-based classifiers,'' \emph{Artificial Intelligence
  Review}, vol.~33, no.~1, pp. 1--39, 2010.

\bibitem{bagging-breiman96}
L.~Breiman, ``Bagging predictors,'' \emph{Machine Learning}, vol.~24, no.~2,
  pp. 123--140, 1996.

\bibitem{breiman-pasting99}
------, ``Pasting small votes for classification in large databases and
  on-line,'' \emph{Machine Learning}, vol.~36, no. 1-2, pp. 85--103, July 1999.

\bibitem{randpeerpodc04}
V.~King and J.~Saia, ``Choosing a random peer,'' in \emph{Proceedings of the
  23rd annual {ACM} symposium on principles of distributed computing
  ({PODC'04})}.\hskip 1em plus 0.5em minus 0.4em\relax ACM Press, 2004, pp.
  125--130.

\bibitem{stutzbach-ton09}
D.~Stutzbach, R.~Rejaie, N.~Duffield, S.~Sen, and W.~Willinger, ``On unbiased
  sampling for unstructured peer-to-peer networks,'' \emph{IEEE/ACM
  Transactions on Networking}, vol.~17, no.~2, pp. 377--390, April 2009.

\bibitem{hall-europar09}
C.~Hall and A.~Carzaniga, ``Uniform sampling for directed {P2P} networks,'' in
  \emph{Euro-Par 2009}, ser. Lecture Notes in Computer Science, H.~Sips,
  D.~Epema, and H.-X. Lin, Eds., vol. 5704.\hskip 1em plus 0.5em minus
  0.4em\relax Springer, 2009, pp. 511--522.

\bibitem{JVGKvS07}
M.~Jelasity, S.~Voulgaris, R.~Guerraoui, A.-M. Kermarrec, and M.~van Steen,
  ``Gossip-based peer sampling,'' \emph{{ACM} Transactions on Computer
  Systems}, vol.~25, no.~3, p.~8, August 2007.

\bibitem{newscastem}
W.~Kowalczyk and N.~Vlassis, ``Newscast {EM},'' in \emph{17th Advances in
  Neural Information Processing Systems ({NIPS})}, L.~K. Saul, Y.~Weiss, and
  L.~Bottou, Eds.\hskip 1em plus 0.5em minus 0.4em\relax Cambridge, MA: MIT
  Press, 2005, pp. 713--720.

\bibitem{p2pdocclust}
S.~Siersdorfer and S.~Sizov, ``Automatic document organization in a p2p
  environment,'' in \emph{Advances in Information Retrieval}, ser. LNCS,
  {Lalmas, M et al.}, Ed.\hskip 1em plus 0.5em minus 0.4em\relax Springer,
  2006, vol. 3936, pp. 265--276.

\bibitem{OHJ10}
R.~Orm{\'a}ndi, I.~Heged{\H{u}}s, and M.~Jelasity, ``Overlay management for
  fully distributed user-based collaborative filtering,'' in \emph{{Euro-Par
  2010}}, ser. Lecture Notes in Computer Science, P.~{D'Ambra}, M.~Guarracino,
  and D.~Talia, Eds., vol. 6271.\hskip 1em plus 0.5em minus 0.4em\relax
  Springer-Verlag, 2010, pp. 446--457.

\bibitem{bakker09}
A.~Bakker, E.~Ogston, and M.~van Steen, ``Collaborative filtering using random
  neighbours in peer-to-peer networks,'' in \emph{Proceeding of the 1st ACM
  international workshop on Complex networks meet information and knowledge
  management ({CNIKM '09})}.\hskip 1em plus 0.5em minus 0.4em\relax New York,
  NY, USA: ACM, 2009, pp. 67--75.

\bibitem{p2prec}
P.~Han, B.~Xie, F.~Yang, J.~Wang, and R.~Shen, ``A novel distributed
  collaborative filtering algorithm and its implementation on p2p overlay
  network,'' in \emph{Advances in Knowledge Discovery and Data Mining}, ser.
  LNCS, H.~Dai, R.~Srikant, and C.~Zhang, Eds.\hskip 1em plus 0.5em minus
  0.4em\relax Springer, 2004, vol. 3056, pp. 106--115.

\bibitem{Tveit-2001}
A.~Tveit, ``Peer-to-peer based recommendations for mobile commerce,'' in
  \emph{Proc. 1st Intl. workshop on Mobile commerce ({WMC '01})}.\hskip 1em
  plus 0.5em minus 0.4em\relax ACM, 2001, pp. 26--29.

\bibitem{luo-kdd07}
P.~Luo, H.~Xiong, K.~L{\"u}, and Z.~Shi, ``Distributed classification in
  peer-to-peer networks,'' in \emph{Proceedings of the 13th ACM SIGKDD
  international conference on Knowledge discovery and data mining
  ({KDD'07})}.\hskip 1em plus 0.5em minus 0.4em\relax New York, NY, USA: ACM,
  2007, pp. 968--976.

\bibitem{ang-ecml09}
H.~Ang, V.~Gopalkrishnan, W.~Ng, and S.~Hoi, ``Communication-efficient
  classification in {P2P} networks,'' in \emph{Machine Learning and Knowledge
  Discovery in Databases ({ECML PKDD})}, ser. Lecture Notes in Computer
  Science, W.~Buntine, M.~Grobelnik, D.~Mladenic, and J.~Shawe-Ta~ylor, Eds.,
  vol. 5781.\hskip 1em plus 0.5em minus 0.4em\relax Springer, 2009, pp. 83--98.

\bibitem{ang-ecml10}
------, ``On classifying drifting concepts in {P2P} networks,'' in
  \emph{Machine Learning and Knowledge Discovery in Databases ({ECML PKDD})},
  ser. Lecture Notes in Computer Science, J.~Balc{\'a}zar, F.~Bonchi,
  A.~Gionis, and M.~Sebag, Eds., vol. 6321.\hskip 1em plus 0.5em minus
  0.4em\relax Springer, 2010, pp. 24--39.

\bibitem{ang-p2pcascade08}
H.~Ang, V.~Gopalkrishnan, S.~Hoi, and W.~Ng, ``Cascade {RSVM} in peer-to-peer
  networks,'' in \emph{Machine Learning and Knowledge Discovery in Databases
  ({ECML PKDD})}, ser. Lecture Notes in Computer Science, W.~Daelemans,
  B.~Goethals, and K.~Morik, Eds., vol. 5211.\hskip 1em plus 0.5em minus
  0.4em\relax Springer, 2008, pp. 55--70.

\bibitem{kargupta-p2pdm06}
S.~Datta, K.~Bhaduri, C.~Giannella, R.~Wolff, and H.~Kargupta, ``Distributed
  data mining in peer-to-peer networks,'' \emph{IEEE Internet Computing},
  vol.~10, no.~4, pp. 18--26, July 2006.

\bibitem{siersdorfer06}
S.~Siersdorfer and S.~Sizov, ``Automatic document organization in a {P2P}
  environment,'' in \emph{Advances in Information Retrieval}, ser. Lecture
  Notes in Computer Science, M.~Lalmas, A.~MacFarlane, S.~R\"uger, A.~Tombros,
  T.~Tsikrika, and A.~Yavlinsky, Eds.\hskip 1em plus 0.5em minus 0.4em\relax
  Springer, 2006, vol. 3936, pp. 265--276.

\bibitem{gadgetsvm09}
C.~Hensel and H.~Dutta, ``{GADGET SVM}: a gossip-based sub-gradient svm
  solver,'' in \emph{International Conference on Machine Learning (ICML),
  Numerical Mathematics in Machine Learning Workshop}, 2009.

\bibitem{adaline-widrow60}
B.~Widrow and M.~E. Hoff, ``Adaptive switching circuits,'' in \emph{1960 {IRE}
  {WESCON} Convention Record, Part 4}.\hskip 1em plus 0.5em minus 0.4em\relax
  New York: {IRE}, 1960, pp. 96--104.

\bibitem{vote99}
E.~Bauer and R.~Kohavi, ``An empirical comparison of voting classification
  algorithms: Bagging, boosting, and variants,'' \emph{Machine Learning},
  vol.~36, no.~1, pp. 105--139, 1999.

\bibitem{peersim}
PeerSim, ``{http://peersim.sourceforge.net/}.''

\bibitem{reuters}
I.~Guyon, A.~B. Hur, S.~Gunn, and G.~Dror, ``Result analysis of the nips 2003
  feature selection challenge,'' in \emph{Advances in Neural Information
  Processing Systems 17}.\hskip 1em plus 0.5em minus 0.4em\relax MIT Press,
  2004, pp. 545--552.

\bibitem{ucirepo}
A.~Frank and A.~Asuncion, ``{UCI} machine learning repository,'' 2010.

\bibitem{p2pchurn}
D.~Stutzbach and R.~Rejaie, ``Understanding churn in peer-to-peer networks,''
  in \emph{Proc. 6th {ACM} Conf. on {Internet} measurement ({IMC'06})}.\hskip
  1em plus 0.5em minus 0.4em\relax ACM, 2006, pp. 189--202.

\bibitem{roozenburg-msc}
J.~Roozenburg, ``Secure decentralized swarm discovery in {Tribler},'' Master's
  thesis, Parallel and Distributed Systems Group, Delft University of
  Technology, 2006.

\bibitem{multiple-forgery-attack}
D.~A. McGrew and S.~R. Fluhrer, ``Multiple forgery attacks against message
  authentication codes,'' \emph{IACR Cryptology ePrint Archive}, vol. 2005, p.
  161, 2005.

\end{thebibliography}
\bibliographystyle{IEEEtran}

\end{document}